\pdfoutput=1

\documentclass[11pt]{article}

\usepackage[final]{acl}

\usepackage{times}
\usepackage{latexsym}
\usepackage{amsmath}
\usepackage[T1]{fontenc}

\usepackage[utf8]{inputenc}

\usepackage{microtype}

\usepackage{inconsolata}

\usepackage{graphicx}
\usepackage{xcolor}         
\usepackage{multirow}
\usepackage{colortbl}
\usepackage{hhline}
\usepackage{tabularx}
\usepackage{subfigure}

\usepackage{wrapfig}
\usepackage{url}
\usepackage[utf8]{inputenc} 
\usepackage[T1]{fontenc}    
\usepackage{hyperref}       
\usepackage{url}            
\usepackage{booktabs}       
\usepackage{amsfonts}       
\usepackage{nicefrac}       
\usepackage{microtype}      

\title{A Bounding Box is Worth One Token - Interleaving Layout and Text in a Large Language Model for Document Understanding}

\author{
  Jinghui Lu\thanks{Equal Contribution \newline \hspace*{0.40cm}\dag Corresponding author}$^{1}$ \ \ Haiyang Yu$^{* 2}$ \ \ Yanjie Wang$^{* 1}$ \ \ Yongjie Ye$^{1}$ \ \ Jingqun Tang$^{1}$ \\
  \textbf{Ziwei Yang$^{1}$ \ \ Binghong Wu$^{1}$ \ \ Qi Liu$^{1}$ \ \ Hao Feng$^{1}$ \ \ Han Wang$^{1}$ \ \ Hao Liu$^{1}$ \ \ Can Huang$^{\dag 1}$}
  \\
  $^1$ByteDance Inc. \ \ $^2$Fudan University\\
  \texttt{lujinghui@bytedance.com, hyyu20@fudan.edu.cn }\\
  \texttt{\{wangyanjie.prince, yeyongjie.ilz, tangjingqun\}@bytedance.com}\\
  \texttt{\{yangziwei.1221, wubinghong, liuqi.nero\}@bytedance.com} \\
  \texttt{\{fenghao.2019, wanghan.99, haoliu.0128, can.huang\}@bytedance.com}}

\begin{document}
\maketitle
\begin{abstract}
Recently, many studies have demonstrated that exclusively incorporating OCR-derived text and spatial layouts with large language models (LLMs) can be highly effective for document understanding tasks. However, existing methods that integrate spatial layouts with text have limitations, such as producing overly long text sequences or failing to fully leverage the autoregressive traits of LLMs. In this work, we introduce \textit{Interleaving \textbf{Lay}out and \textbf{Text} in a \textbf{L}arge \textbf{L}anguage \textbf{M}odel (LayTextLLM)} for document understanding. LayTextLLM projects each bounding box to a single embedding and interleaves it with text, efficiently avoiding long sequence issues while leveraging autoregressive traits of LLMs. LayTextLLM not only streamlines the interaction of layout and textual data but also shows enhanced performance in KIE and VQA. Comprehensive benchmark evaluations reveal significant improvements of LayTextLLM, with a 15.2\% increase on KIE tasks and 10.7\% on VQA tasks compared to previous SOTA OCR-based LLMs. All resources are available at \url{https://github.com/LayTextLLM/LayTextLLM}.

\end{abstract}

\section{Introduction}


Recent research has increasingly explored the use of Large Language Models (LLMs) or MultiModal Large Language Models (MLLMs)~\citep{gpt4v,team2023gemini,anthropic2024claude,reid2024gemini,feng2023docpedia,feng2023unidoc,liu2024textmonkey,lu2024padellmner,nourbakhsh-etal-2024-towards,gao-etal-2024-ttm,li-etal-2024-hypergraph,zhou-etal-2024-llms,zhu-etal-2024-fanoutqa,zhao-etal-2024-docmath} for document-oriented Visual Question Answering (VQA) and Key Information Extraction (KIE). 






A line of research utilizes off-the-shelf OCR tools to extract text and spatial layouts, which are then combined with LLMs to address Visually Rich Document Understanding (VRDU) tasks. These approaches assume that \textit{most valuable information for document comprehension can be derived from the text and its spatial layouts, viewing spatial layouts as ``lightweight visual information''~\citep{wang-etal-2024-docllm}}. Following this premise, several studies~\citep{liu2024textmonkey,perot2023lmdx,luo2024layoutllm,chen2023shikra,he2023icl} have explored various approaches that integrate spatial layouts with text for LLMs and achieves results that are competitive with those of MLLMs. 

The most natural method to incorporate layout information is by treating spatial layouts as tokens, which allows for the seamless interleaving of text and layout into a unified text sequence~\citep{perot2023lmdx,chen2023shikra,he2023icl}. For example, ~\citet{perot2023lmdx} employ format such as \textit{``HARRISBURG 78|09''} to represent OCR text and corresponding layout, where \textit{``HARRISBURG''} is OCR text and \textit{``78|09''} indicates the mean of the horizontal and vertical coordinates, respectively. Similarly, ~\citet{he2023icl} use \textit{``[x\_min, y\_min, x\_max, y\_max]''} to represent layout information. These approaches can effectively take advantage of autoregressive characteristics of LLMs and is known as the \textit{``coordinate-as-tokens''} scheme~\citep{perot2023lmdx}. In contrast, DocLLM~\citep{wang-etal-2024-docllm} explores interacting spatial layouts with text through a disentangled spatial attention mechanism that captures cross-alignment between text and layout modalities. 

\begin{figure}[t]  
  \centering
  \includegraphics[width=0.45\textwidth]{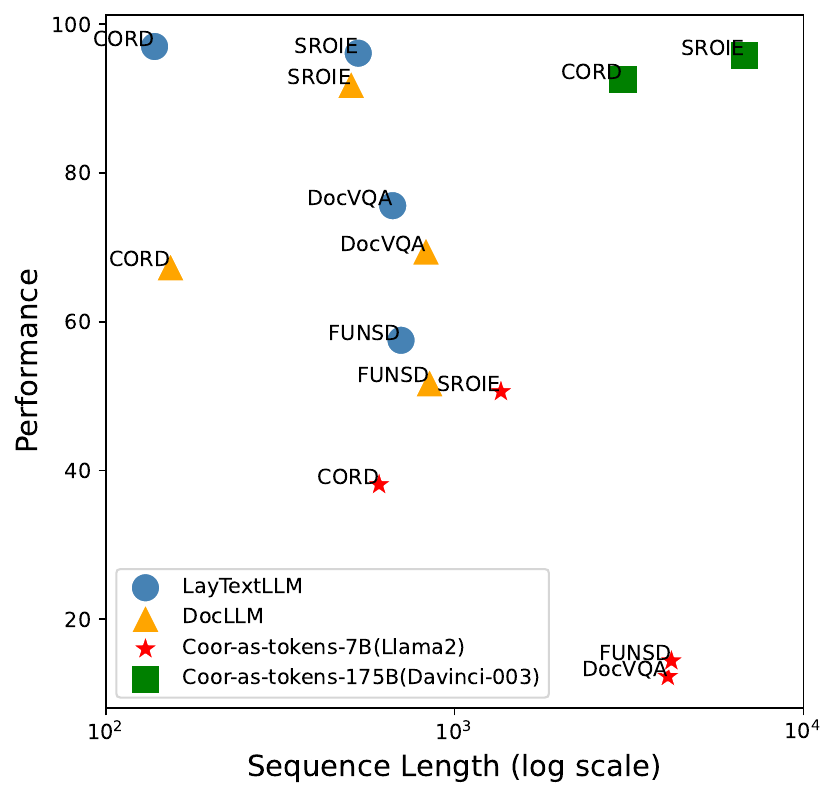}
  \caption{The performance against input sequence length of different datasets across various OCR-based methods where data is from Table~\ref{tab:docllm} and~\ref{tab:length}.}
  \label{fig:perf_vs_length}
\end{figure}

However, we believe that both of the previous approaches have limitations. As shown in Figure~\ref{fig:perf_vs_length}, coordinate-as-tokens significantly increases the number of tokens. Additionally, to accurately comprehend coordinates and enhance zero-shot capabilities, this scheme often requires few-shot in-context demonstrations and large-scale language models, such as ChatGPT Davinci-003 (175B)~\citep{he2023icl}, which exacerbates issues related to sequence length and GPU resource demands. Although DocLLM does not increase sequence length, its performance may be improved by more effectively leveraging the autoregressive traits of LLMs.


To address these problems, this paper explores a simple yet effective approach to enhance the interaction between spatial layouts and text --- \textit{Interleaving
\textbf{Lay}out and \textbf{Text} in a \textbf{L}arge \textbf{L}anguage \textbf{M}odel (LayTextLLM)} for document understanding. Adhering to the common practice of interleaving any modality with text~\citep{kosmos,kosmos2,dong2024internlm2}, we specifically apply this principle to spatial layouts. In particular, we map each bounding box to a single embedding, which is then interleaved with its corresponding text. As shown in Figure~\ref{fig:perf_vs_length}, LayTextLLM significantly outperforms the 175B models, while only slightly increasing or even reducing the sequence length compared to DocLLM. Our contributions can be listed as follows:



\begin{itemize}
    \item We propose LayTextLLM for document understanding. To the best of the authors' knowledge, this is the first work to employ a unified embedding approach (Section~\ref{subsec:model_arch}) that interleaves spatial layouts directly with textual data within a LLM. By representing each bounding box with one token, LayTextLLM efficiently addresses sequence length issues brought by coordiante-as-tokens while fully leveraging autoregressive traits for VRDU tasks.
    \item We propose three tailored pre-training tasks (Section~\ref{subsec:pre-train}) to improve the model's understanding of the interaction between layout and text, and its ability to generate precise coordinates for regions of interest. These tasks include Line-level Layout Decoding, Text-to-Layout Prediction, and Layout-to-Text Prediction. Besides, we introduce Spatially-Grounded KIE (Section~\ref{subsec:sft}) to further enhance the model’s performance on KIE task.
    \item Extensive experimental results quantitatively demonstrate that LayTextLLM significantly surpasses previous state-of-the-art (SOTA) OCR-based methods. Notably, it outperforms DocLLM by 10.7\% on VQA tasks and 15.2\% on KIE tasks (Section~\ref{sec:exp}). Furthermore, it achieves superior performance on SOTA OCR-free MLLMs, such as Qwen2-VL among most KIE datasets. Ablations and visualizations demonstrate the utility of the proposed component, with analysis showing that LayTextLLM not only improves performance but also reduces input sequence length compared to current OCR-based models.
\end{itemize}

\section{Related Work}

\subsection{OCR-based LLMs for VRDU}



Early document understanding methods~\citep{Hwang_Yim_Park_Yang_Seo_2020,Xu_Li_Cui_Huang_Wei_Zhou_2020,LayoutLMV2,Hong_Kim_Ji_Hwang_Nam_Park_2022,Tang_Yang_Wang_Fang_Liu_C_Zeng_Cha_Bansal_2022} tend to solve the task in a two-stage manner, \textit{i.e.}, first reading texts from input document images using off-the-shelf OCR engines and then understanding the extracted texts. Considering the advantages of LLMs (\textit{e.g.}, high generalizability), some recent methods endeavor to combine LLMs with OCR-derived results to solve document understanding. Inspired by the coordinate-as-tokens'' approach in ICL-D3IE~\citep{perot2023lmdx}, \citet{he2023icl} use numerical tokens to integrate layout information, combining layout and text into a unified sequence that maximizes the autoregressive benefits of LLMs. To reinforce the layout information while avoiding increasing the number of tokens, DocLLM~\citep{wang-etal-2024-docllm} designs a disentangled spatial attention mechanism to capture cross-alignment between text and layout modalities. Recently, LayoutLLM~\citep{luo2024layoutllm} utilizes the pre-trained layout-aware model~\citep{huang2022layoutlmv3}, to insert the visual information, layout information and text information. However, these methods struggle to leverage autoregressive properties of LLMs while avoiding the computational overhead of increasing token counts. Finding a way to integrate layout information remains a challenge.

\subsection{OCR-free MLLMs for VRDU}

With the increasing popularity of MLLMs~\citep{feng2023unidoc,hu2024mplug,liu2024textmonkey,tang2024textsquare,chen2024far,dong2024internlm2,li2024mini,liu2024llavanext}, various methods are proposed to solve VRDU through explicitly training models on visual text understanding datasets and perform end-to-end inference without using OCR engines. LLaVAR~\citep{zhang2023llavar} and UniDoc~\citep{feng2023unidoc} are notable examples that expand upon the document-oriented VQA capabilities of LLaVA~\citep{liu2024visual} by incorporating document-based tasks. These models pioneer the use of MLLMs for predicting texts and coordinates from document images, enabling the development of OCR-free document understanding methods. Additionally, DocPedia~\citep{feng2023docpedia} operates document images in the frequency domain, allowing for higher input resolution without increasing the input sequence length. Recent advancements in this field, including mPLUG-DocOwl~\citep{mPLUG-DocOwl}, Qwen-VL~\citep{bai2023qwen}, Qwen2-VL~\citep{wang2024qwen2}, and TextMonkey~\citep{liu2024textmonkey}, leverage publicly available document-related VQA datasets to further enhance the document understanding capability. Although these OCR-free methods have exhibited their advantages, they still struggle with the high-resolution input to reserve more text-related details.

\section{Method}\label{sec:method}

\begin{figure*}[t]
	\centering
	\includegraphics[width=0.9\textwidth]{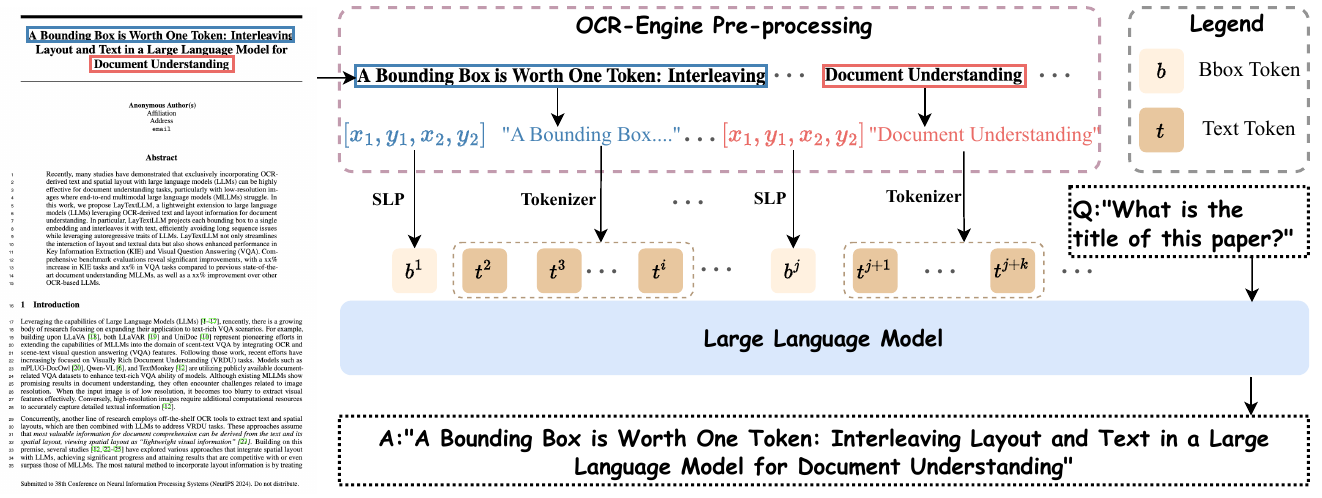}
	\caption{An overview of LayTextLLM incorporates interleaving bounding box tokens ($b^i$) with text tokens ($t^i$), where the superscripts represent the sequence positions of the tokens.}
	\label{fig:overview}
\end{figure*}

\begin{figure*}[ht]
    \centering
    \subfigure[Line-level Layout Decoding]{        \vtop{\vskip0pt\hbox{\includegraphics[width=0.48\textwidth]{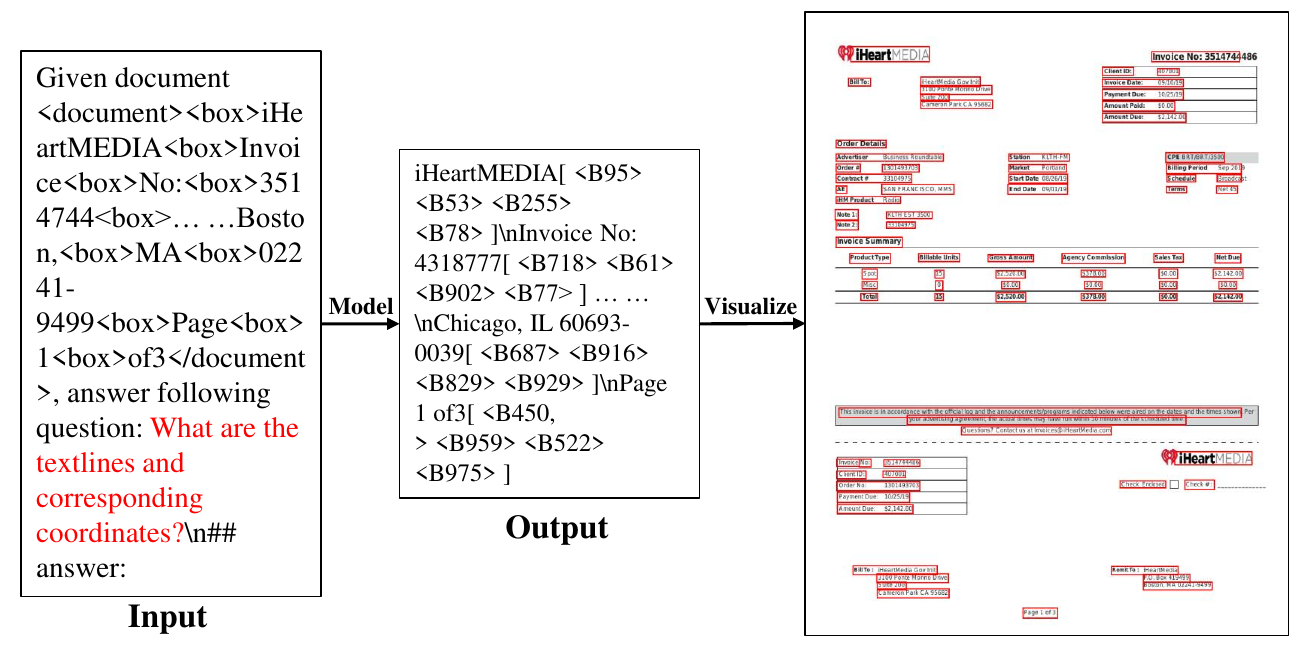}}}
       
    }
    \subfigure[Text-to-layout Prediction]{        \vtop{\vskip0pt\hbox{\includegraphics[width=0.48\textwidth]{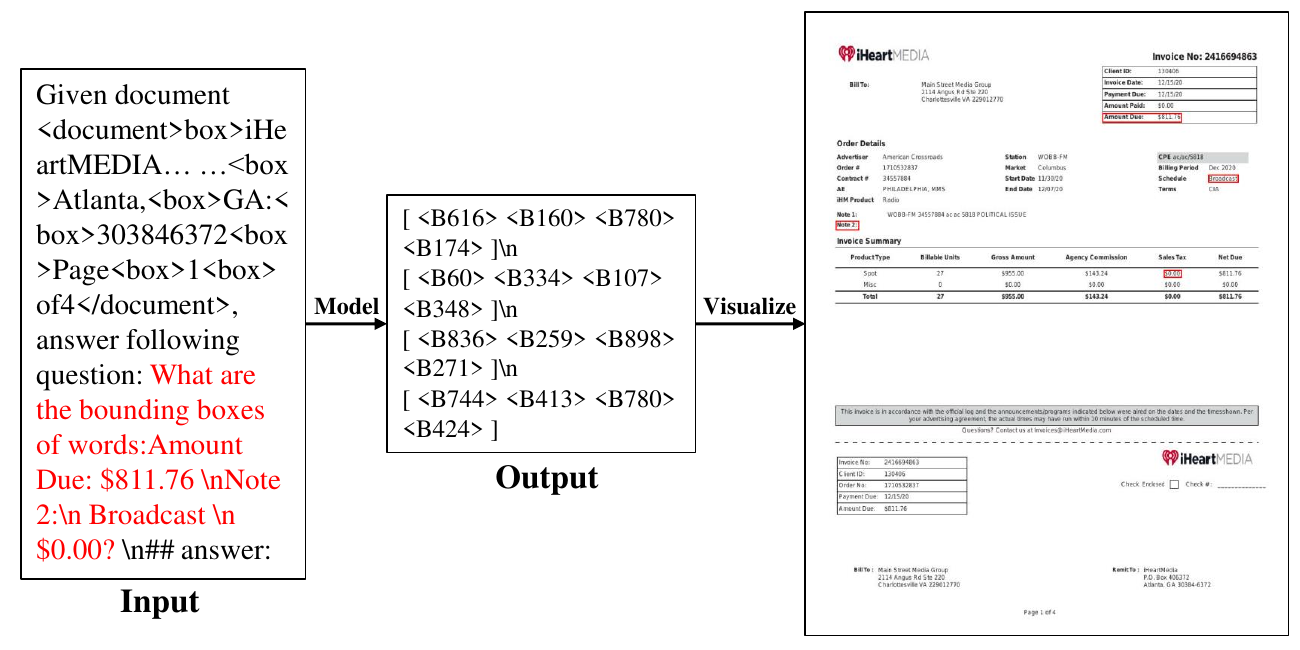}}}
       
    }
    \subfigure[Layout-to-text Prediction]{ \vtop{\vskip0pt\hbox{\includegraphics[width=0.48\textwidth]{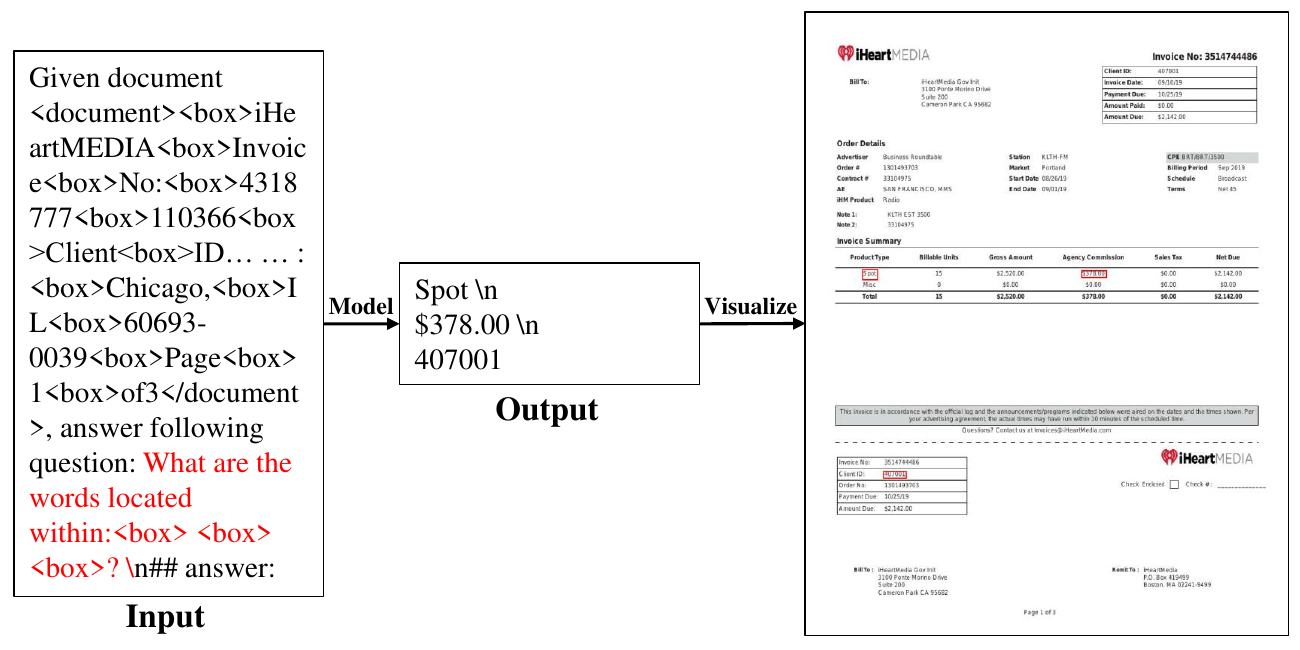}}}
       
    }
    \caption{Illustration of layout-text alignment pre-training tasks. <box> is the placeholder for bounding box tokens.}
    \label{fig:tasks}
\end{figure*}

\begin{figure*}[ht]
    \centering
    \subfigure[SG-KIE for Entity Linking]{        \vtop{\vskip0pt\hbox{\includegraphics[width=0.48\textwidth]{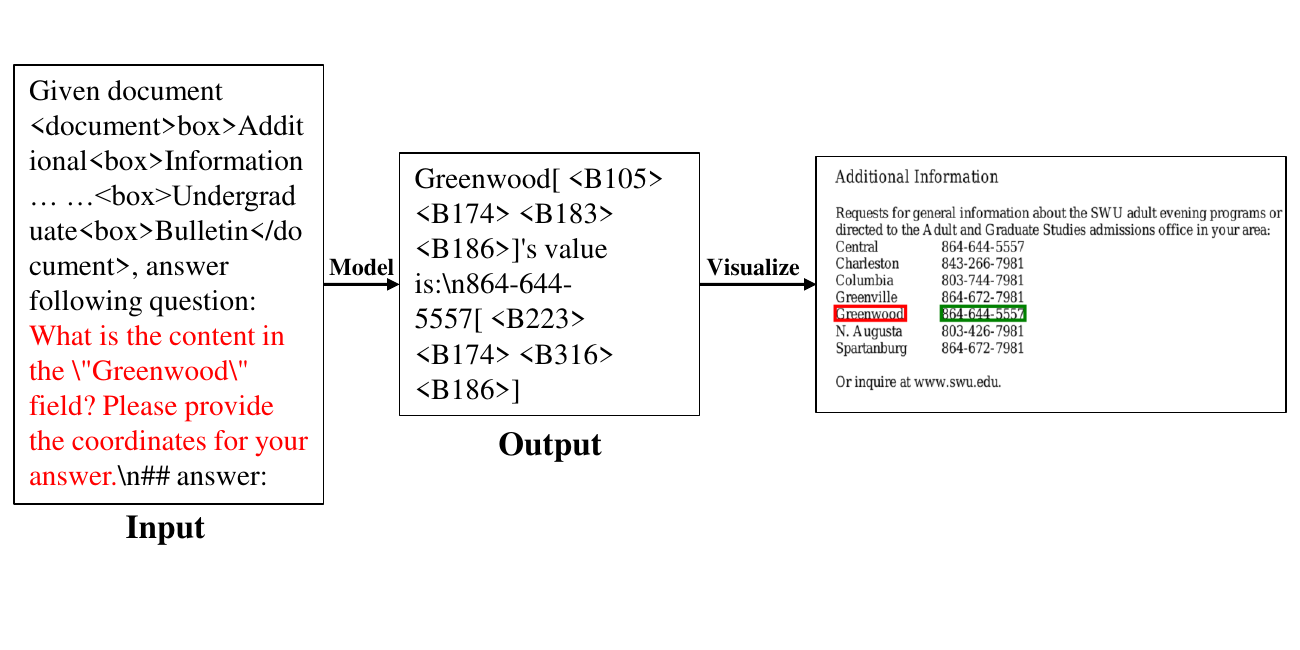}}}
       
    }
    \subfigure[SG-KIE for Semantic Entity Recognition]{        \vtop{\vskip0pt\hbox{\includegraphics[width=0.48\textwidth]{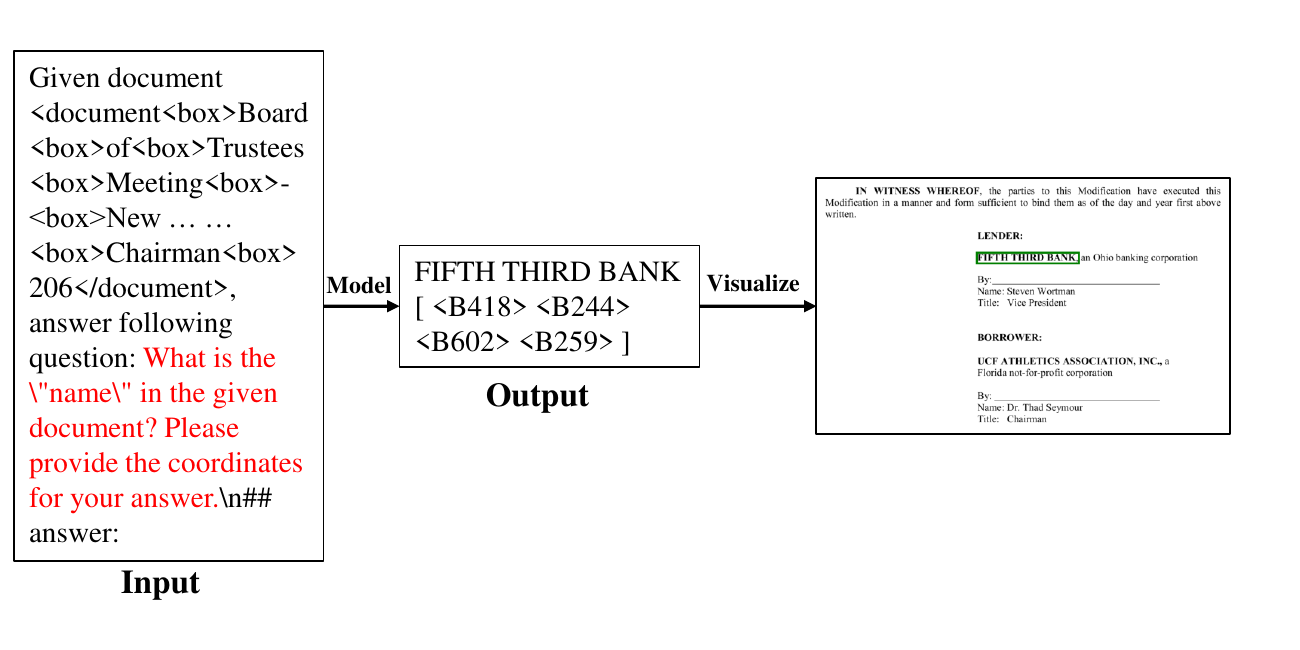}}}
       
    }
    \caption{Illustration of Spatially-Grounded KIE task. <box> is the placeholder for bounding box tokens.}
    \label{fig:sg_kie_tasks}
\end{figure*}


In this section, we introduce LayTextLLM. We begin by detailing the model architecture, which features an innovative Spatial Layout Projector (Section~\ref{subsec:model_arch}) that transforms four-dimensional layout coordinates into a single-token embedding. Next, we present three layout-text alignment pre-training tasks: line-level layout decoding, text-to-layout prediction, and layout-to-text prediction (Section~\ref{subsec:pre-train}) to ensure a seamless integration of layout and text understanding. Finally, we describe the incorporation of spatially-grounded key information extraction as a auxiliary task during supervised fine-tuning (SFT) (Section~\ref{subsec:sft}), to enhance the performance in KIE tasks.

\subsection{Model Architecture}\label{subsec:model_arch}

The overall architecture of LayTextLLM is shown in Figure~\ref{fig:overview}. LayTextLLM is built on the Llama2-7B-chat model~\citep{gao2023llama}.

\paragraph{Spatial Layout Projector} To enable the model to seamlessly integrate spatial layouts with text, we propose a novel \textbf{S}patial \textbf{L}ayout \textbf{P}rojector (SLP). This projector employs a two-layer MLP to transform layout coordinates into bounding box tokens, facilitating the interleaving of spatial and textual information. Concretely, each OCR-derived spatial layout is represented by a bounding box defined by four-dimensional coordinates $[x_{1},y_{1},x_{2},y_{2}]$, where these coordinates denote the normalized minimum and maximum horizontal ($x$) and vertical ($y$) extents of the box, respectively. The SLP maps these coordinates into a high-dimensional embedding space, enabling the LLM to process them as a single token. This is computed as:
\begin{equation}
    z = W_{2} \cdot (\text{GeLU}(W_{1} \cdot c + b_{1})) + b_{2}
\end{equation} 

\noindent where \( c \in \mathbb{R}^{4} \) is the vector of bounding box coordinates, \( W_{1} \in \mathbb{R}^{h \times 4} \) and \( W_{2} \in \mathbb{R}^{d \times h} \) are weight matrices, \( b_{1} \in \mathbb{R}^{h \times 1} \) and \( b_{2} \in \mathbb{R}^{d \times 1} \) are bias vectors, \( h \) is the hidden dimension of the MLP, and \( d \) is the dimension of the final embedding. In this study, we set $h=d$. The resulting bounding box token \( z \in \mathbb{R}^{d} \) is a high-dimensional representation of the spatial layout. Importantly, the SLP is shared across all bounding box tokens, which introduces a minimal number of parameters to the model.


\paragraph{Large Language Model} As shown in Figure~\ref{fig:overview}, the bounding box token 
$z$ is interleaved with its corresponding textual embeddings and fed into the LLM. To introduce additional trainable parameters for layout information, we integrate a Partial Low-Rank Adaptation (P-LoRA) module proposed in InternLM-XComposer2~\citep{dong2024internlm2} detailed in Appendix~\ref{appendix:p_lora}. Additionally, to improve the efficiency of coordinate decoding, we introduce 1,000 special tokens, \textit{i.e.,} \textit{``<B0>''} through \textit{``<B999>''} to represent output coordinates. 


\subsection{Training Tasks}\label{sec:training}

LayTextLLM is pre-trained using three innovative tasks designed to align layout and text. During the SFT phase, we introduce a novel Spatially-Grounded Key Information Extraction task as a auxiliary task, which significantly enhances the model's performance on KIE-related tasks. Figures~\ref{fig:tasks} and~\ref{fig:sg_kie_tasks} illustrate the above tasks.

\subsubsection{Layout-text Alignment Pre-training}\label{subsec:pre-train}

\paragraph{Line-level Layout Decoding}

To enhance the model's ability to interpret and reconstruct layout information, we introduce the Line-level Layout Decoding task. This task leverages the bounding box embeddings, which encode spatial layout details, and challenges the model to decode these embeddings back into precise coordinates. Specifically, the model is provided with word-level OCR texts and their corresponding layout coordinates as input. It is then prompted with the question: \textit{``What are the textlines and corresponding coordinates?''} The model is expected to intelligently merge word-level OCR texts into coherent line-level texts while simultaneously generating the coordinates that represent the layout of these line-level texts. The output consists of two components: (1) the reconstructed line-level texts and (2) the corresponding combined coordinates, which are derived by aggregating the word-level bounding boxes to reflect the spatial arrangement of the line-level OCR. Through this task, the model is expected to demonstrate two key abilities: (1) the ability to logically group word-level texts into line-level texts using layout information, and (2) the ability to accurately decode bounding box embeddings back into spatial coordinates. By doing so, the model demonstrates a deeper understanding of both textual content and its spatial organization within a document.





\paragraph{Text-to-layout Prediction}

To enhance the model's ability to comprehend and predict document layouts, we introduce the Text-to-Layout Prediction task. In this task, the model predicts spatial coordinates for text segments based on word-level OCR inputs and their corresponding layout information. Specifically, given a prompt such as \textit{``What are the bounding boxes of the words: \{word1\} \textbackslash n \{word2\}  \textbackslash n \{word3\}...?''}, where \{word\} represents line-level text randomly selected from the input (number of selected words limited to 5), the model is required to generate precise spatial coordinates for each of the specified words. 

\paragraph{Layout-to-text Prediction}

We also propose the Layout-to-Text Prediction task. In this task, the model predicts textual content based on spatial layout information and bounding box coordinates. Given a prompt such as \textit{``What are the words located within: \{bbox1\} \textbackslash n \{bbox2\} \textbackslash n \{bbox3\}...?''}, where \{bbox\} is the placeholder of bounding box embedding representing the spatial coordinates of text regions (with the number of bounding boxes limited to 5), the model generates the corresponding textual content for each specified region. The Text-to-Layout Prediction and Layout-to-Text Prediction tasks offer complementary advantages to advance document layout understanding. All word-level and line-level OCR results can be easily obtained using off-the-shelf OCR tools, making it easy to scale up for large-scale pre-training.

\subsubsection{Supervised Fine-tuning}\label{subsec:sft}

During the SFT phase, we fine-tuned the pre-trained model with the Document Dense Description (DDD) and Layout-aware SFT datasets from~\citet{luo2024layoutllm}. Additionally, we introduce \textbf{S}patially-\textbf{G}rounded \textbf{K}ey \textbf{I}nformation \textbf{E}xtraction (SG-KIE) task, which requires the model to not only answer questions (\textit{i.e.,} extract specific values) but also provide the coordinates of these answers by responding to the prompt \textit{``Please provide the coordinates for your answer.''} as a auxiliary task to further improve the model performance on KIE tasks.

In the literature, KIE tasks are classified into two types: Entity Linking (EL) and Semantic Entity Recognition (SER). EL is an open-set KIE task in which both the key and its corresponding value are present in the input. In contrast, SER is a closed-set KIE task where the key has a predefined meaning, and the value must be extracted from the document.

For the EL task, SG-KIE requires the model to output the answer in the following format: 
\textit{``\{key\}\{key\_bbox\}'s value is \{value\}\{value\_bbox\}''}, 
where \{key\} and \{value\} represent the respective key and value, and \{key\_bbox\} and \{value\_bbox\} denotes the spatial layout information of the corresponding textual content. For the SER task, the answer format is:
\textit{``\{value\}\{value\_bbox\}''}, 
where \{value\} refers to the extracted value, and \{value\_bbox\} represents the spatial layout of the extracted text in the document. The illustrations of SG-KIE for these tasks are presented in Figure~\ref{fig:sg_kie_tasks}.

\section{Experiments}\label{sec:exp}

\subsection{Datasets}
\label{subsec:dataset}

\noindent\textbf{Layout-text Alignment Pre-training Data} \ \ \ In training process, we exclusively used open-source data to facilitate replication. We subsampled data from two datasets for layout-text alignment pre-training: (1) \textbf{DocILE}~\citep{vsimsa2023docile} and (2) \textbf{RVL\_CDIP}~\citep{harley2015evaluation}. 

\noindent\textbf{SFT data} \ \ \ We selected \textbf{KVP10k}~\citep{naparstek2024kvp10k} and \textbf{SIBR}~\citep{yang2023modeling} datasets to create training examples of SG-KIE tasks. For document-oriented VQA, we selected \textbf{Document Dense Description (DDD)} and \textbf{Layout-aware SFT} data used in~\citet{luo2024layoutllm}, which are two synthetic datasets generated by GPT-4. Besides, \textbf{DocVQA}~\citep{mathew2021docvqa}, \textbf{InfoVQA}~\citep{mathew2022infographicvqa}, \textbf{ChartQA}~\citep{masry-etal-2022-chartqa}, \textbf{VisualMRC}~\citep{tanaka2021visualmrc} is included following~\citep{liu2024textmonkey}. For KIE task, we selected \textbf{SROIE}~\citep{huang2019icdar2019}, \textbf{CORD}~\citep{park2019cord}, \textbf{FUNSD}~\citep{jaume2019funsd} datasets following~\citet{wang-etal-2024-docllm,luo2024layoutllm,liu2024textmonkey}. The dataset statistics are provided in Appendix~\ref{appendix:data}.

\subsection{Implementation Detail}
\label{subsec:implementation}

The LLM component of LayTextLLM is initialized from the Llama2-7B-chat~\citep{touvron2023llama}, consistent with previous OCR-based methods like DocLLM~\citep{wang-etal-2024-docllm}, which also use Llama2-7B. We also replicated the results of the coor-as-tokens scheme using Llama2-7B for consistency. Noting the LayoutLLM~\citep{luo2024layoutllm} utilizes Llama2-7B and Vicuna 1.5 7B, which is fine-tuned from Llama2-7B. Thus, for the majority of our comparisons, the models are based on the same or similar LLM backbones, allowing for a fair comparison between approaches. Other MLLM baselines use backbones like Qwen-VL~\citep{bai2023qwen}, Qwen2-VL~\citep{wang2024qwen2}, InternVL~\citep{chen2024internvl}, and Vicuna~\citep{chen2024far}, all with at least 7B parameters, excluding the visual encoder. This also makes the comparison fair.

In this study, we developed two versions of LayTextLLM to facilitate a comparative analysis under different training configurations. Following the terminology established by~\citet{luo2024layoutllm}, the term ``zero-shot'' denotes models that are trained without exposure to data from downstream test datasets. For the first version, \textbf{LayTextLLM$_{zero}$}, we utilized DDD, Layout-aware SFT data, KVP10k, and SIBR for training. The second version, \textbf{LayTextLLM$_{all}$}, extends this training regimen by incorporating a broader array of VQA and KIE datasets, including DocVQA, InfoVQA, VisualMRC, ChartQA, FUNSD, CORD, and SROIE. Both versions are initialized with the same pre-trained LayTextLLM weights, with the key difference being that LayTextLLM$_{all}$ benefits from the inclusion of additional downstream training datasets. We used word-level and line-level OCR provided by the respective datasets for a fair comparison, with the exception of the ChartQA dataset, which does not provide OCR. Detailed setup can be found in Appendix~\ref{appendix:imple}.

\subsection{Baselines}

\noindent\textbf{OCR-based baselines} \  \  \   For OCR-based baseline models, we implemented a basic approach using only OCR-derived text as input. This was done using two versions: \textbf{Llama2-7B-base} and \textbf{Llama2-7B-chat}. We also adapted the coordinate-as-tokens scheme from~\citet{he2023icl} for these models, resulting in two new variants: \textbf{Llama2-7B-base$_{coor}$} and \textbf{Llama2-7B-chat$_{coor}$}. Additionally, we included results from a stronger baseline using the ChatGPT Davinci-003 (175B) model~\citep{he2023icl}, termed \textbf{Davinci-003-175B$_{coor}$}. One other recent SOTA OCR-based approach, \textbf{DocLLM}~\citep{wang-etal-2024-docllm} is also included. 

\noindent\textbf{OCR-free baselines}  \  \  \  These baselines include \textbf{UniDoc}~\citep{feng2023unidoc}, \textbf{DocPedia}~\citep{feng2023docpedia}, \textbf{Monkey}~\citep{li2023monkey}, \textbf{InternVL}~\citep{chen2023internvl}, \textbf{InternLM-XComposer2}~\citep{dong2024internlm2}, \textbf{TextMonkey}, \textbf{TextMonkey$_{+}$}~\citep{liu2024textmonkey}, \textbf{Qwen2-VL}~\citep{wang2024qwen2}. We selected the above models as baselines due to their superior performance in both document-oriented VQA and KIE tasks.

\noindent\textbf{Visual+OCR baselines} \  \  \ We selected \textbf{LayoutLLM$_{Llama2^{CoT}}$}~\citep{luo2024layoutllm} and the most recent SOTA method \textbf{DocLayLLM$_{Llama2^{CoT}}$}~\citep{liao2024doclayllm}, which integrates visual cues, text and layout, as stronger baselines.

\subsection{Evaluation Metrics}

To ensure a fair comparison with other OCR-based methods, we conducted additional evaluations using original metrics specific to certain datasets, such as F1 score~\citep{wang-etal-2024-docllm,he2023icl}, ANLS~\citep{Gao_ICDAR_2019,wang-etal-2024-docllm,luo2024layoutllm} and CIDEr~\citep{vedantam2015cider,wang-etal-2024-docllm}.
To ensure a fair comparison with OCR-free methods, we adopted the accuracy metric~\citep{liu2024textmonkey,feng2023unidoc}, where a response from the model is considered correct if it fully captures the ground truth. 

\begin{table*}[t]
    \small
    \centering
    \renewcommand\arraystretch{1.0}
    \scalebox{0.75}{
    \begin{tabular}{l|ccc|cccc}
    \toprule    
     ~ & \multicolumn{3}{c|}{\textbf{Document-Oriented VQA}} & \multicolumn{4}{c}{\textbf{KIE}} \\
     ~ & DocVQA & VisualMRC & Avg & FUNSD & CORD & SROIE & Avg \\
    \midrule
    \textbf{Metric} & \multicolumn{3}{c|}{\textit{ANLS \% / CIDEr}} & \multicolumn{4}{c}{\textit{F-score \%}} \\
    \midrule
    \textbf{Text} & & & & & & & \\
    Llama2-7B-base & 34.0 & 182.7 & 108.3 & 25.6 & 51.9 & 43.4 & 40.3 \\
    Llama2-7B-chat & 20.5 & 6.3 & 13.4 & 23.4 & 51.8 & 58.6 & 44.6 \\
    \midrule
    \textbf{Text + Coordinates} & & & & & & &\\
    Llama2-7B-base$_{coor}$~\citep{he2023icl} & 8.4 & 3.8 & 6.1 & 6.0 & 46.4 & 34.7 & 29.0 \\
    Llama2-7B-chat$_{coor}$~\citep{he2023icl} & 12.3 & 28.0 & 20.1 & 14.4 & 38.1 & 50.6 & 34.3 \\
    Davinci-003-175B$_{coor}$~\citep{he2023icl} & - & - & - & - & 92.6 & 95.8 & -\\
    DocLLM~\citep{wang-etal-2024-docllm} & 69.5$^{*}$ & 264.1$^{*}$ & 166.8 & 51.8$^{*}$ & 67.4$^{*}$ & 91.9$^{*}$ & 70.4 \\\midrule
    LayTextLLM$_{zero}$ (Ours) & 66.6 & 229.1 & 147.9 & 57.6 & 87.3 & 89.4 & 78.1 \\
    LayTextLLM$_{all}$ (Ours) & \textbf{75.6$^{*}$} & \textbf{279.4$^{*}$} & \textbf{177.5} & \textbf{63.3$^{*}$} & \textbf{97.3$^{*}$} & \textbf{96.0$^{*}$}  & \textbf{85.6} \\
    \bottomrule
    \end{tabular}
    }
        \caption{Comparison with SOTA OCR-based methods. The asterisk(*) indicates that the model was trained using the training set associated with the evaluation set.}
    \label{tab:docllm}
\end{table*}

\begin{table*}[t]
    \small
    \centering
    \scalebox{0.75}{
    \begin{tabular}{l|ccc|ccccc}
    \toprule    
     ~ & \multicolumn{3}{c|}{\textbf{Document-Oriented VQA}} & \multicolumn{5}{c}{\textbf{KIE}} \\
     ~ & DocVQA & InfoVQA & Avg & FUNSD & SROIE & POIE & CORD & Avg \\
    \midrule
    \textbf{Metric} & \multicolumn{8}{c}{\textit{Accuracy \%} } \\
    \midrule
    \textbf{OCR-free} & & & & & & & &\\
    UniDoc~\citep{feng2023unidoc} & 7.7 & 14.7  & 11.2 & 1.0 & 2.9  & 5.1 & -& - \\
    DocPedia~\citep{feng2023docpedia} & 47.1$^{*}$ & 15.2$^{*}$  & 31.2& 29.9 & 21.4 & 39.9 & -& - \\
    Monkey~\citep{li2023monkey} & 50.1$^{*}$ & 25.8$^{*}$  & 38.0 &24.1 & 41.9 & 19.9 & -& - \\
    InternVL~\citep{chen2023internvl} & 28.7$^{*}$ & 23.6$^{*}$  & 26.2 & 6.5 & 26.4 & 25.9 & -& -\\
    InternLM-XComposer2~\citep{dong2024internlm2} & 39.7 & 28.6  & 34.2 & 15.3 & 34.2 & 49.3 & -& -\\
    TextMonkey~\citep{liu2024textmonkey} & 64.3$^{*}$ & 28.2$^{*}$  & 46.3 & 32.3 & 47.0 & 27.9 & -& -\\
    TextMonkey$_{+}$~\citep{liu2024textmonkey} & 66.7$^{*}$ & 28.6$^{*}$  & 47.7 & 42.9 & 46.2 & 32.0 & -& -\\
    Qwen2-VL~\citep{wang2024qwen2} & \textbf{81.4$^{*}$} & \textbf{45.2$^{*}$}  & \textbf{63.3}  & 53.2 & 71.3 & \textbf{85.7} & 78.8 & 72.2\\
    \midrule
    \textbf{Text + Coordinates} & & & &  & & \\
    LayTextLLM$_{zero}$ (Ours) & 70.4 & 29.8 & 50.1 & 54.9 & 88.3 &  65.1 &  86.9 & 73.8\\
    LayTextLLM$_{all}$ (Ours) & 77.7$^{*}$ & 40.1$^{*}$ &  59.0 & \textbf{60.1$^{*}$} & \textbf{95.5$^{*}$} &   68.1 & \textbf{96.7$^{*}$}& \textbf{80.1}  \\
    \bottomrule
    \end{tabular}
    }
        \caption{Comparison with SOTA OCR-free MLLMs.}
    \label{tab:main_cmp}
\end{table*}

\begin{table*}[t]
    \small
    \centering
    \renewcommand\arraystretch{0.95}
    \scalebox{0.75}{
    \begin{tabular}{l|ccc|cccc}
    \toprule    
     ~ & \multicolumn{3}{c|}{\textbf{Document-Oriented VQA}} & \multicolumn{4}{c}{\textbf{KIE}} \\
     ~ & DocVQA & VisualMRC & Avg & FUNSD$^{-}$ & CORD$^{-}$ & SROIE$^{-}$ & Avg \\
    \midrule
    \textbf{Metric} & \multicolumn{7}{c}{\textit{ANLS \%}} \\
    \midrule
    \textbf{Visual + Text + Coordinates} & & & & \\
    LayoutLLM$_{Llama2^{CoT}}$~\citep{luo2024layoutllm} & 74.2 & \textbf{55.7} & \textbf{64.9} & 78.6 & 62.2 & 70.9 & 70.6 \\
    DocLayLLM$_{Llama2^{CoT}}$~\citep{liao2024doclayllm} & 72.8 & 55.0 & 63.9 & 78.7 & 70.8 & 83.2 & 77.6 \\
    \midrule
    \textbf{Text + Coordinates} & & & & & & &\\
    
    LayTextLLM$_{zero}$ (Ours) & 66.6 & 37.9 & 52.3 & 79.0 & 79.8 & 90.2  & 83.0 \\
    LayTextLLM$_{all}$ (Ours) & \textbf{75.6$^{*}$} & 42.3$^{*}$ & 59.0 & \textbf{83.4$^{*}$} & \textbf{83.1$^{*}$} &  \textbf{95.6$^{*}$} & \textbf{87.4}  \\
    \bottomrule

    \end{tabular}
    }
        \caption{Comparison with LayoutLLM. The superscript minus($^{-}$) indicates that the cleaned test set used in~\citet{luo2024layoutllm}.}
    \label{tab:layoutllm}
\end{table*}

\begin{table*}[t]
\centering
\scalebox{0.60}{\begin{tabular}{cccc|cccc|cccc}
\toprule
~ &~&~& ~&\multicolumn{4}{c|}{\textbf{Document-Oriented VQA}} & \multicolumn{4}{c}{\textbf{KIE}} \\\midrule
 SLP & L-T PT & SG-KIE & P-LoRA  & DocVQA & InfoVQA & VisualMRC  & Avg &FUNSD & CORD & SROIE & Avg  \\    \midrule
$\times$ & \checkmark & \checkmark  & \checkmark & 65.8 & 25.3 & 28.7 & 39.9 & 49.3 & 65.8 &  61.9 & 59.0 \\
\checkmark & $\times$ & \checkmark  & \checkmark & \textbf{78.2} & \textbf{39.7} & 28.3 & \textbf{48.7} & 52.1 & 76.5 & 86.8 & 71.8 \\
\checkmark & \checkmark & $\times$  & \checkmark & 69.1 & 28.7 & 29.3 &  42.3&52.3 & 82.4  & 84.0 & 72.9 \\
\checkmark & \checkmark & \checkmark &  $\times$ & 74.6 & 36.6  & \textbf{32.6} & 47.9 & 54.8 & 86.0 & \textbf{91.3} & \textbf{77.4}  \\
\checkmark & \checkmark & \checkmark & \checkmark & 70.4 & 29.8  &31.7 & 44.0 & \textbf{54.9} & \textbf{86.9} &88.3 & 76.7  \\

\bottomrule
\end{tabular}
}
\caption{Ablations on each component of LayTextLLM (Accuracy).}
\label{tab:ablation}
\end{table*}

\subsection{Quantitative Results}

\noindent\textbf{Comparison with SOTA OCR-based Methods} \ \ \ For the primary comparison in our work, we evaluate against other SOTA pure OCR-based methods. The experimental results, as presented in Table~\ref{tab:docllm}, demonstrate significant performance improvements achieved by the LayTextLLM models compared to DocLLM~\citep{wang-etal-2024-docllm}. Specifically, LayTextLLM$_{zero}$ exhibits notably superior performance, with its zero-shot capabilities even rivaling SFT approaches. For instance, in the KIE task, LayTextLLM$_{zero}$ achieves an overall performance of 78.1\%, significantly outperforming DocLLM's score of 70.4\%. Furthermore, under the same training conditions, LayTextLLM$_{all}$ surpasses the previous OCR-based SOTA by a substantial margin, achieving an overall improvement of 10.7\% in the VQA task and 15.2\% in the KIE tasks. Besides, we found that the spatial information can be decoded back into coordinates even without visual information, as discussed in Appendix~\ref{appendix:decode_box}, which is not exhibited in DocLLM. Similarly, when contrasting with coordinate-as-tokens employed in Llama2-7B, LayTextLLM$_{zero}$ again outperforms significantly. More qualitative results are shown in Appendix~\ref{appendix:quali}. More discussion of subperformance of DocLLM and coordinate-as-tokens can be seen Appendix~\ref{appendix:interleave}.






\noindent\textbf{Comparison with SOTA OCR-free Methods} \ \ \ We also compare LayTextLLM with other OCR-free methods, and the results in Table~\ref{tab:main_cmp} highlight its exceptional performance across various tasks. Due to fairness concerns, results for ChartQA are reported separately in Appendix~\ref{appendix:chartqa}, as the dataset lacks OCR-derived outputs, and we employed in-house OCR tools instead.

LayTextLLM$_{zero}$ significantly outperforms most OCR-free methods except for Qwen2-VL. Notably, even without exposure to the dataset’s training set, LayTextLLM$_{zero}$ achieves competitive VQA performance, rivaling models like TextMonkey${+}$, which were trained on corresponding datasets. When fine-tuned with relevant data, LayTextLLM$_{all}$ exhibits even greater performance improvements. Compared to the SOTA MLLM Qwen2-VL, LayTextLLM sub-performs on VQA tasks which is further discussed in Limitation (Section~\ref{sec:limit}). However, it outperforms Qwen2-VL in terms of KIE tasks. Notably, LayTextLLM$_{zero}$ exceeds Qwen2-VL on three out of four KIE benchmarks, with significant improvements of 1.7\% on FUNSD, 17\% on SROIE, and 8.1\% on CORD.

\noindent\textbf{Comparison with SOTA Visual+OCR Methods} \ \ \ As shown in Table~\ref{tab:layoutllm}, in zero-shot scenarios, our approach outperforms LayoutLLM and DocLayLLM on most KIE datasets, with improvements of 12.4\% and 5.4\%, respectively. This is noteworthy given that both LayoutLLM and DocLayLLM utilize visual, OCR text, and layout information as inputs and inference with Chain-of-thought, highlighting our ability to effectively leverage OCR-based results. However, similar to the comparison results with MLLMs, LayTextLLM exhibits limitations in document-oriented VQA tasks, particularly when addressing questions that heavily depend on visual information. A more detailed analysis of these challenges is provided in Limitations (Section~\ref{sec:limit}).

\subsection{Analysis}\label{sec:analysis}





\noindent\textbf{Ablations} \ \ \ To better assess the utility of each component in LayTextLLM, an ablation study was conducted, the results of which are presented in Table~\ref{tab:ablation}. Detailed information on the training setup for all variants is provided in Appendix~\ref{appendix:imple}. The results clearly show that incorporating interleaved spatial layouts and texts significantly enhances the performance, evidenced by a  4.1\% improvement in VQA and a 17.7\% increase in KIE (the first row vs. the fourth row), indicating that SLP is a critical component. Interestingly, using next-token-prediction as the pre-training task (\textit{i.e.,} the second row) generally outperforms layout-text alignment pre-training across almost all VQA tasks. However, for KIE tasks, layout-text alignment pre-training remains more effective. We hypothesize that layout-text alignment pre-training helps the model learn the relationship between layout and text, which is particularly useful for layout-aware tasks like KIE. In contrast, next-token-prediction focuses on reconstructing the entire document, which is more beneficial for semantic-rich tasks like VQA. Furthermore, including SG-KIE results in a modest performance increase of 1.7\% in VQA (the third row vs. the fourth row) but a significant improvement in KIE tasks (\textit{i.e.,} 3.8\%), which is as expected. Intriguingly, excluding P-LoRA improves performance on VQA and KIE tasks, suggesting it adds unnecessary complexity or interference, which further highlights the benefits of interleaving texts and layouts.


\noindent\textbf{Sequence Length} \ \ \ Table~\ref{tab:length} presents statistics on the average input sequence length across different datasets. Intriguingly, despite interleaving bounding box tokens, LayTextLLM consistently exhibits the shortest sequence length in three out of four datasets, even surpassing DocLLM, which is counterintuitive. We attribute this to the tokenizer mechanism. For example, using \texttt{tokenizer.encode()}, a single word from the OCR engine, like \textit{``International''} is encoded into a single ID $[4623]$. Conversely, when the entire OCR output is processed as one sequence, such as \textit{``... CPC,International,Inc...''}, the word \textit{``International''} is split into two IDs $[17579, 1288]$, corresponding to \textit{``Intern''} and \textit{``ational''} respectively. This type of case occurs frequently, we provide further discussion in Appendix~\ref{appendix:length}. 


\begin{table}[h]
\centering
\scalebox{0.65}{\begin{tabular}{c|ccc}
\toprule
Dataset & LayTextLLM & DocLLM & Coor-as-tokens \\
\midrule
DocVQA & \textbf{664.3} & 827.5 &  4085.7\\
CORD & \textbf{137.9} & 153.2 & 607.3 \\
FUNSD & \textbf{701.9} & 847.5 & 4183.4  \\
SROIE & 529.2 & \textbf{505.1} & 1357.7 \\
\bottomrule
\end{tabular}}
\caption{Average sequence length.}\label{tab:length}
\end{table}
\vspace{-7pt}

\section{Conclusion}\label{sec:conclu}

We propose LayTextLLM, interleaving spatial layouts and text to improve predictions through an innovative SLP, the Layout-text Alignment pre-training and the SG-KIE tasks. Extensive experiments show the effectiveness of LayTextLLM.

\section*{Limitations}\label{sec:limit}

Although LayTextLLM has shown significant capabilities in text-rich VQA and KIE tasks, this alone does not suffice for all real-world applications. There are some instances where reasoning must be based solely on visual cues (\textit{e.g.} size, color, objects)—a challenge that remains unmet. Questions such as \textit{``What is the difference between the highest and the lowest green bar?''} and \textit{``What is written on the card on the palm?''} illustrate this gap. Two bad cases, detailed in Figures~\ref{fig:failure_case_charqa} and~\ref{fig:failure_case_docvqa}, also underscore these limitations. Addressing these challenges underscores the need for future advancements that incorporate visual cues into the capabilities of LayTextLLM. Since the integration with MLLMs is not the primary focus of this work, the preliminary experiments exploring this approach are discussed in Appendix~\ref{appendix:mllm}.

\bibliography{acl_latex}
\newpage
\twocolumn
\appendix

\section{Layout Partial Low-Rank Adaptation}\label{appendix:p_lora}

\begin{figure}[t]  
  \centering
  \includegraphics[width=1.0\columnwidth]{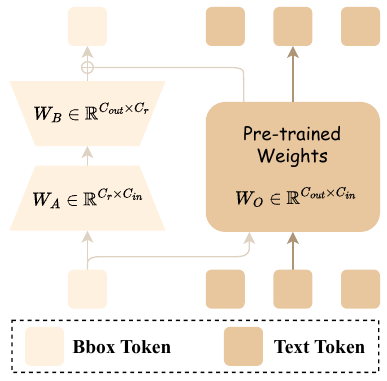}
  \caption{The illustration of P-LoRA, adapted from~\citet{dong2024internlm2}.}
  \label{fig:lora}
\end{figure}

\begin{figure*}[ht]
	\centering
	\includegraphics[width=0.8\textwidth]{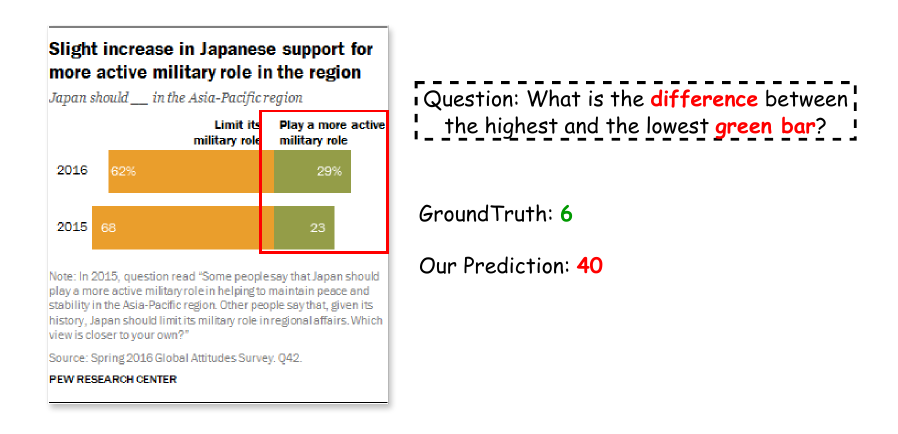}
	\caption{A failure case of LayTextLLM on ChartQA.}
	\label{fig:failure_case_charqa}
\end{figure*}

\begin{figure*}[ht]
	\centering
	\includegraphics[width=0.8\textwidth]{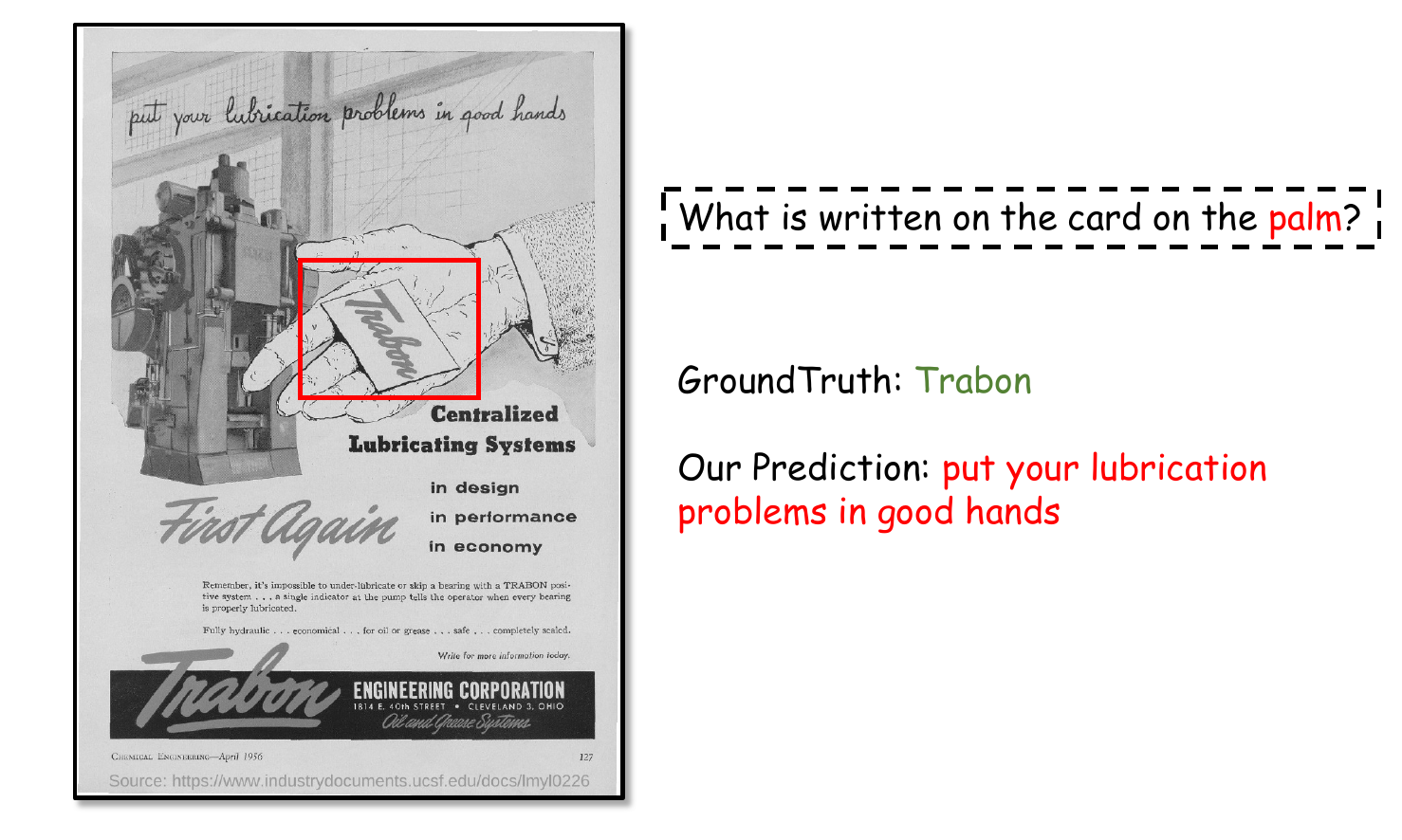}
	\caption{A failure case of LayTextLLM on DocVQA.}
	\label{fig:failure_case_docvqa}
\end{figure*}

After using the SLP to generate bounding box tokens and a tokenizer to produce text tokens, these two modalities are then interacted using a Layout Partial Low-Rank Adaptation (P-LoRA) module in LLMs. P-LoRA, introduced in InternLM-XComposer2~\citep{dong2024internlm2}, is originally used to adapt LLMs to the visual modality. It applies plug-in low-rank modules specified to the visual tokens, which adds minimal parameters while preserving the LLMs inherent knowledge.

Formally, for a linear layer in the LLM, the original weights $W_{O} \in \mathbb{R}^{C_{out} \times C_{in}}$ and bias $B_{O} \in \mathbb{R}^{C_{out}}$ are specified for input and output dimensions $C_{in}$ and $C_{out}$. P-LoRA modifies this setup by incorporating two additional matrices, $W_{A} \in \mathbb{R}^{C_{r} \times C_{in}}$ and $W_{B} \in \mathbb{R}^{C_{out} \times C_{r}}$. These matrices are lower-rank, with $C_{r}$ being considerably smaller than both $C_{in}$ and $C_{out}$, and are specifically designed to interact with new modality tokens, which in our case are bounding box tokens. For example, given an input $x=[x_{b}, x_{t}]$ comprising of bounding box tokens ($x_{b}$) and textual tokens ($x_{t}$) is fed into the system, the forward process is as follows, where $\hat{x}_t, \hat{x}_b$ and $\hat{x}$ are outputs:

\begin{equation}
\begin{aligned}
& \hat{x}_t=W_0 x_t+B_0 \\
& \hat{x}_b=W_0 x_b+W_B W_A x_b+B_0 \\
& \hat{x}=\left[\hat{x}_b, \hat{x}_t\right]
\end{aligned}
\end{equation}

\section{Qualitative Examples}\label{appendix:quali}

\begin{figure*}[ht]
	\centering
	\includegraphics[width=1.0\textwidth]{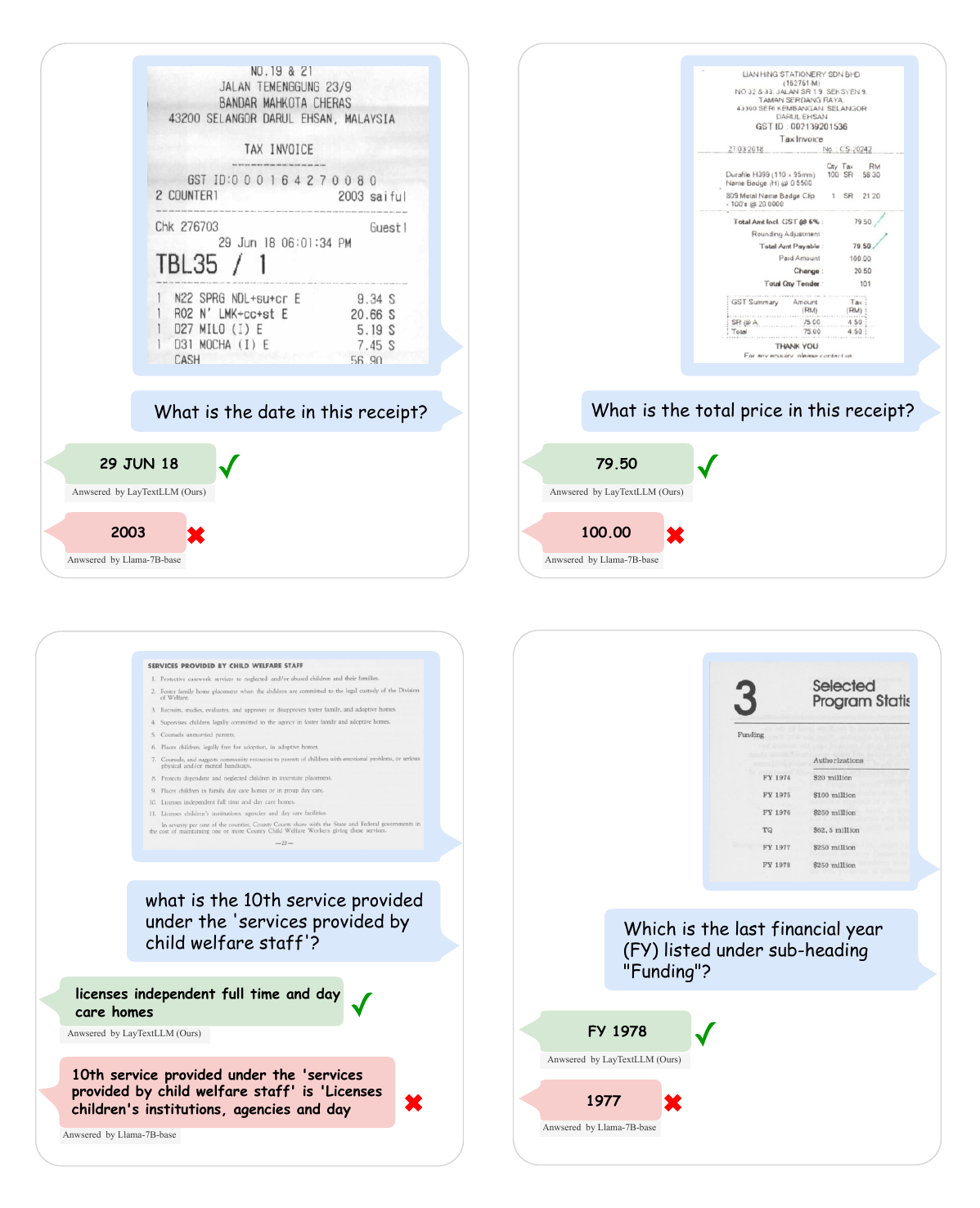}
	\caption{Qualitative comparison with the baseline method.}
	\label{fig:qualitative_examples_KIE}
\end{figure*}

Qualitative examples of document-oriented VQA (upper row) and KIE (bottom row) are shown in Figure~\ref{fig:qualitative_examples_KIE}. The results indicate that LayTextLLM is highly effective in utilizing spatial layout information to make more accurate predictions for these challenging examples. For example, in the upper right figure, many numeric texts in the receipt act as noise for the baseline method. In contrast, LayTextLLM integrates layout information to accurately predict the total price, as demonstrated by the other examples, underscoring the utility of LayTextLLM.

\section{Dataset Statistics}\label{appendix:data}

Table~\ref{tab:stats_pt} and~\ref{tab:stats_sft} show the statistics of datasets used in layout-text alignment pre-training and SFT, respectively. In layout-text alignment pre-training, for training efficiency, we randomly selected around 50,000 documents from each of the DocILE and RVL\_CDIP datasets. For every document, we generated two tasks: line-level layout decoding and either a text-to-layout or layout-to-text prediction task, which yields a total of around 200,000 pre-training examples. We also tested the model on a KIE dataset \textbf{POIE}~\citep{kuang2023visual}.

\begin{table}[ht]
\centering
\scalebox{0.8}{\begin{tabular}{c|cc}
\toprule
Dataset & DocILE & RVL\_CDIP  \\
\midrule
Num Documents & 55,719 & 59444  \\
Num Examples & 111,438 & 118,888  \\
Num Tokens & 75,952,078 & 67,340,246   \\
\bottomrule
\end{tabular}}
\caption{Dataset statistics for layout-text alignment pre-training (using Llama-2 Tokenizer).}\label{tab:stats_pt}
\end{table}

\begin{table*}[ht]
\centering
\scalebox{0.6}{\begin{tabular}{c|cccccccccccc}
\toprule
Dataset & DDD & Layout-aware SFT & KVP10k & SIBR & DocVQA & InfoVQA & ChartQA & VisualMRC & FUNSD & CORD & SROIE \\
\midrule
Num Documents & 115,955 & 50,409&4,249&600&10,192&4,405&3,699&7,012& 147 &794&626 \\
Num Examples & 115,955 & 280,073&50,661&12,978&39,459&23,945&7,398&7,013&2,375&8,932&2,503 \\
Num Tokens & 71,067,212 & 101,209,393&27,018,563&8,045,694&17,621,621&1,024,236&1,052,752&1,622,387&11,543,711
&1,140,437&1,066,930 \\
\bottomrule
\end{tabular}}
\caption{Dataset statistics for SFT (using Llama-2 Tokenizer).}\label{tab:stats_sft}
\end{table*}

\section{Implementation Detail}\label{appendix:imple}

All training and inference procedures are conducted on eight NVIDIA A100 GPUs.

\paragraph{Training} LayTextLLM is initialized with Llama2-7b-chat model, the pre-training, SFT, and other model hyper-parameters can be seen in Table~\ref{tab:setup}. Additional parameters including SLP and P-LoRA are randomly initialized. During pre-training and SFT, all parameters are trainable. Please note that all variants of LayTextLLM, including those utilized in ablation studies, are trained in accordance with the same settings. Specifically, for all variants in ablation study, we train with the same setting and dataset in accordance with LayTextLLM$_{zero}$. For the variant without SLP, we replace the bounding box token placeholder \textit{``<box>''} with \textit{``\textbackslash n''}. For the variant without layout-text alignment pre-training, we pre-train the model on the same dataset using a conventional next-token prediction task, excluding the loss computation for the bounding box token. After pre-training, we fine-tune the model on the SFT datasets. For the variant without SG-KIE tasks, we remove the SG-KIE data from the SFT datasets while retaining the original SER and EL tasks in KVP10k and SIBR to ensure the total number of training examples remains unchanged. For the variant without P-LoRA, we replace all P-LoRA modules with linear layers, as was previously done.

All baseline results are sourced from \citet{liu2024textmonkey} or respective original papers, with the exception of the Llama2-7B series, the Llama2-7B$_{\text{coor}}$ series, and Qwen2-VL, these results were re-implemented by authors.

\begin{table*}[ht]
\centering
\scalebox{0.75}{\begin{tabular}{@{}lllllllll@{}}
\toprule
& \textbf{Backbone} & \textbf{Plora rank} & \textbf{Batch size} & \textbf{Max length} & \textbf{Precision} & \textbf{Train params} & \textbf{Fix params}  \\ 
\midrule
\textbf{Lay-Text Pretrain} & Llama2-7B-base & 256 & 128 & 4096 & bf16 & 7.4 B & 0B  \\
\textbf{SFT}      & Llama2-7B-base & 256 & 128 & 4096 & bf16 & 7.4 B & 0B  \\
\midrule
& \textbf{Learning rate} & \textbf{Weight decay} & \textbf{Scheduler} & \textbf{Adam betas} & \textbf{Adam epsilon} & \textbf{Warm up} &  \textbf{Epoch}& \\
\midrule
\textbf{Lay-Text Pretrain} & 5.0e-05 & 0.01 & cosine & [0.9, 0.999] & 1.0e-08 & 0.005 &4  \\
\textbf{SFT}      & 1.0e-05 & 0.01 & cosine & [0.9, 0.999] & 1.0e-08 & 0.005 &4  \\
\bottomrule
\end{tabular}}
\caption{LayTextLLM trainng Hyper-parameters.}\label{tab:setup}
\end{table*}

\paragraph{Inference} 

For the document-oriented VQA test set, we use the original question-answer pairs as the prompt and ground truth, respectively. For KIE tasks, we reformat the key-value pairs into a question-answer format, as described in~\citet{wang-etal-2024-docllm, luo2024layoutllm, liu2024textmonkey}. Additionally, for the FUNSD dataset, we focus our testing on the entity linking annotations as described in~\citet{luo2024layoutllm}. Note that for KIE tasks, we report the result of directly generating the answer texts, instead of generating the answer with the coordinates (SG-KIE). The discussion regarding inference with SG-KIE can be found in Appendix~\ref{appendix:sg_kie}.

To eliminate the impact of randomness on evaluation, no sampling methods are employed during testing for any of the models. Instead, beam search with a beam size of 1 is used for generation across all models. Additionally, the maximum number of new tokens is set to 512, while the maximum number of input tokens is set to 4096.

\section{Discussion of Input Sequence Length}\label{appendix:length}

As mentioned in Section \ref{sec:analysis}, it is intriguing that LayTextLLM has fewer input sequences than DocLLM, which is counterintuitive given that LayTextLLM interleaves bounding box tokens, typically resulting in longer sequence lengths. We attribute this to the Byte Pair Encoding (BPE) tokenizers~\citep{bpe} prevalently used in modern LLMs such as Llama2.

BPE operates by building a vocabulary of commonly occurring subwords (or token pieces) derived from the training data. Initially, it tokenizes the text at the character level and then progressively merges the most frequent adjacent pairs of characters or sequences. The objective is to strike a balance between minimizing vocabulary size and maximizing encoding efficiency.

Thus, when tokenizing a single word like \textit{``International''} on its own, the tokenizer might identify it as a common sequence in the training data and encode it as a single token. This is especially likely if \textit{``International''} frequently appears as a standalone word in the training contexts. However, when the word \textit{``International''} is part of a larger sequence of words such as including in a long sequence of OCR-derived texts like \textit{``...335
CPC,International,Inc...''}, the context changes. The tokenizer might split \textit{``International''} into sub-tokens like \textit{``Intern''} and \textit{``ational''} because, in various contexts within the training data, these subwords might appear more frequently in different combinations or are more useful for the model to understand variations in meaning or syntax. 

When using LayTextLLM, we input word-level OCR results into the tokenizer, typically resulting in the former situation, where words are encoded as single tokens. Conversely, with DocLLM, the entire OCR output is processed as one large sequence, leading to the latter situation and a longer sequence length than in LayTextLLM. This difference underscores the utility of LayTextLLM in achieving both accuracy and inference efficiency due to its shorter sequence length.

\section{Discussion on Advantage of Interleaving Layout and Text}\label{appendix:interleave}

\begin{figure*}[ht]
    \centering
    \subfigure[Attention map of the first layer.]{        \vtop{\vskip0pt\hbox{\includegraphics[width=0.31\textwidth]{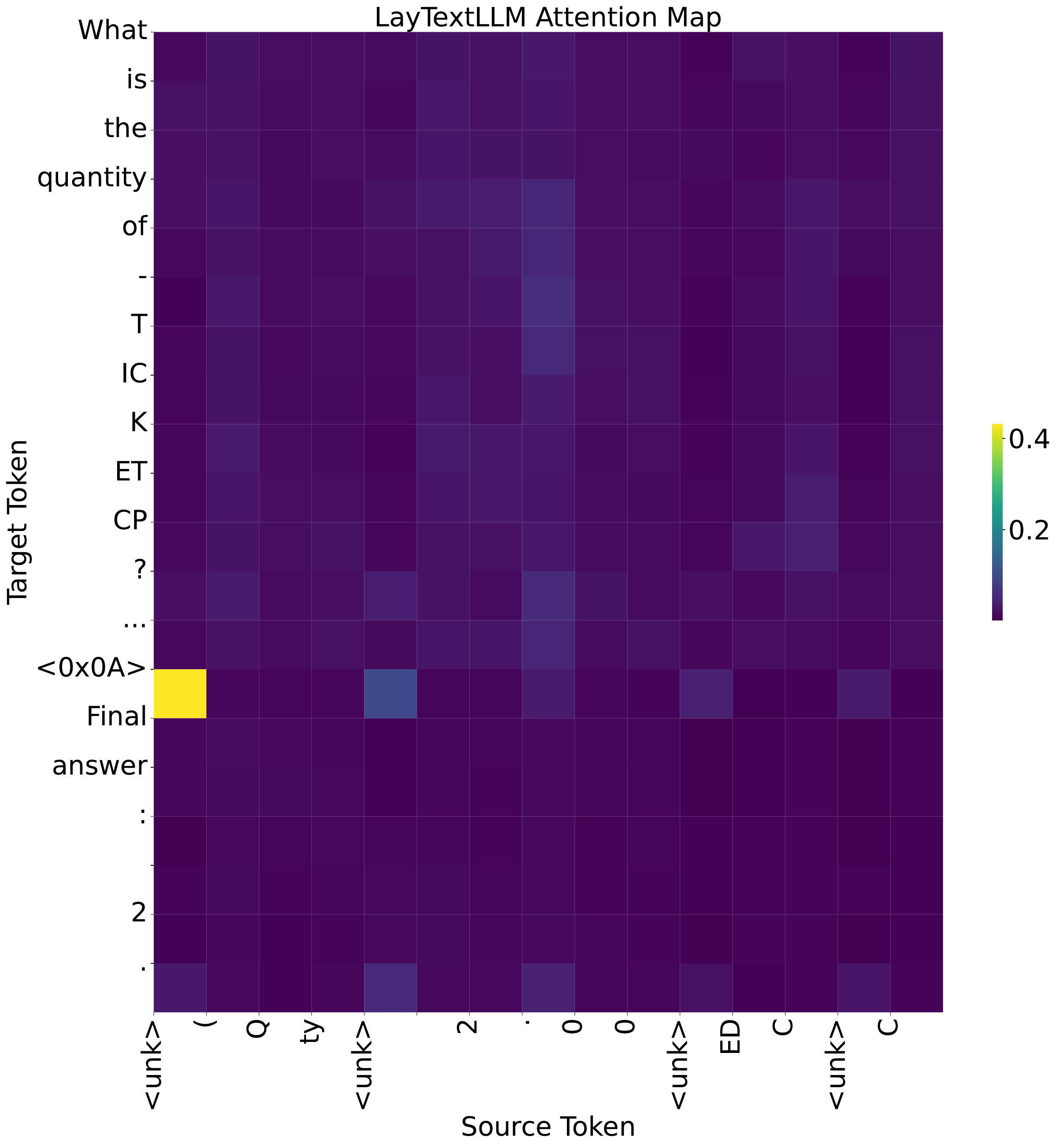}}}
        \label{fig:attention_first_layer}
    }
    \subfigure[Attention map of the last layer.]{        \vtop{\vskip0pt\hbox{\includegraphics[width=0.32\textwidth]{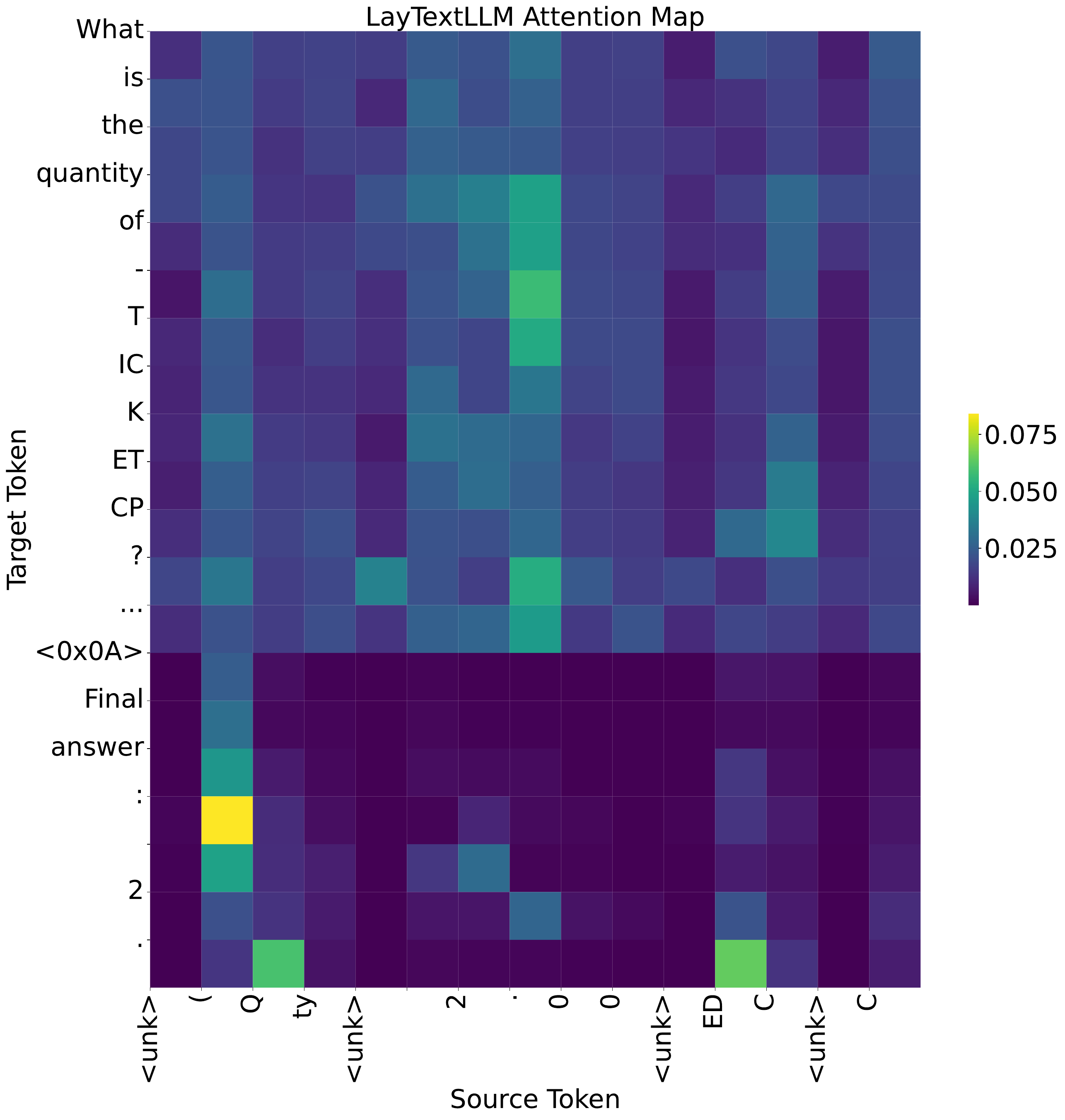}}}
        \label{fig:attention_last_layer}
    }
    \subfigure[Attention score visualization of bounding box tokens.]{ \vtop{\vskip0pt\hbox{\includegraphics[width=0.31\textwidth]{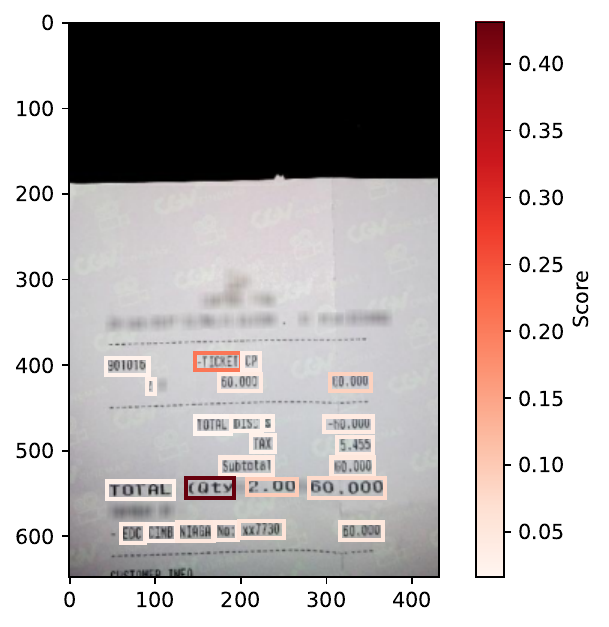}}}
        \label{fig:attention_map3}
    }
    \caption{Visualization of attention maps of LayTextLLM. Best viewed in color and with zoom. \textit{``<unk>''} is the placeholder for the bounding box token.}
    \label{fig:attention_maps}
\end{figure*}

\paragraph{Discussion on DocLLM} We visualize the attention patterns between input and output tokens in Figure~\ref{fig:attention_maps}. The attention pattern is insightful with the specific question, \textit{``What is the quantity of - TICKET CP?<0x0A>''}

As shown in Figure~\ref{fig:attention_first_layer}, when the model begins predicting the answer \textit{``Final''}, \textit{``<0x0A>''(newline symbol)} is heavily focusing on layout information, as seen by the significant attention on the bounding box embedding \textit{``<unk>''} token before \textit{``(Qty''}. This highlights the model's effort to orient itself spatially and understand the structural context of the tokens. At this stage, the model is developing a cognitive understanding of how the elements are laid out on the page. We extract and visualize the attention scores that \textit{``<0x0A>''} assigns to each bounding box in Figure~\ref{fig:attention_map3}. The visualization shows that the model focuses most on \textit{``Qty''}, followed by \textit{``-TICKET''} and \textit{``2.00''}, which reflects the layout information essential for making the prediction. In the final layer (Figure~\ref{fig:attention_last_layer}), the model's attention shifts dramatically towards the \textit{``Qty''} token, which holds the semantic meaning necessary to answer the question. This shift from layout-based cognition to content-based reasoning illustrates how the bounding box tokens act as spatial anchors that help the model pinpoint and organize the relevant information (such as \textit{``Qty''}) to make the correct prediction.

The attention of LayTextLLM exhibits a distinct pattern compared to models like DocLLM, which uses block infilling to predict missing blocks from both preceding and succeeding context. In contrast, LayTextLLM adheres to an auto-regressive approach, focusing its attention solely on preceding information. Furthermore, interleaving bounding box and text embeddings creates strong attention connections between textual and spatial representations, as shown in Figure~\ref{fig:attention_maps}. In contrast, DocLLM integrates spatial information into the calculation of attention score which is implicitly. As shown in Table~\ref{tab:docllm}, LayTextLLM significantly outperforms DocLLM, again underscoring the advantage of interleaving bounding box and text embeddings. Also, we found that the spatial information can be decoded back into coordinates even without inputing visual information, as discussed in Appendix~\ref{appendix:decode_box}, which is not exhibited in DocLLM.

We also conduct a fairer experiment by re-implementing DocLLM using the identical training settings as LayTextLLM$_{zero}$. In order to ensure a more intuitive and fair comparison between the two layout adaptation methods (\textit{i.e.,} SLP versus disentangled spatial attention), we exclude the use of P-LoRA in LayTextLLM$_{zero}$. Table~\ref{tab:slp_vs_attention} demonstrates that SLP is a more effective method for incorporating layout information, as evidenced by a 6.7\% improvement in VQA and an 8.4\% improvement in KIE. Additionally, while DocLLM introduces a suite of attention weights for layout information, it significantly increases the number of parameters in LLaMA-2 from 6.73B to 8.37B. In contrast, LayTextLLM introduces a much smaller increase in parameters.

\begin{table*}[t]
\centering
\scalebox{0.7}{\begin{tabular}{c|cccc|cccc|c}
\toprule
~ &\multicolumn{4}{c|}{\textbf{Document-Oriented VQA}} & \multicolumn{4}{c|}{\textbf{KIE}} & \textbf{Num Params} \\\midrule
 Methods & DocVQA & InfoVQA & VisualMRC  & Avg &FUNSD & CORD & SROIE & Avg  \\    \midrule
 DocLLM  & 66.6& 28.3 & 28.6 & 41.2 & 51.3 & 71.8 &  83.9 & 69.0 &8.37B \\
LayTextLLM & \textbf{74.6} & \textbf{36.6}  & \textbf{32.6} & \textbf{47.9} &\textbf{54.8} & \textbf{86.0} & \textbf{91.3} &  \textbf{77.4} & \textbf{6.76B} \\

\bottomrule
\end{tabular}
}
\caption{Comparison of two layout adaptation methods, \textit{i.e.,} SLP in LayTextLLM and Disentangled Spatial Attention in DocLLM.}
\label{tab:slp_vs_attention}
\end{table*}

\paragraph{Discussion on coordinate-as-tokens} The subperformance of coordinate-as-tokens methods can be attributed to the following three reasons: (1) The coordinate-as-tokens approach tends to introduce an excessive number of tokens, often exceeding the pre-defined maximum length of Llama2-7B (\textit{i.e.}, 4096). Consequently, this leads to a lack of crucial OCR information, resulting in hallucination and subpar performance. (2) When re-implementing the coordinate-as-tokens method with Llama2-7B, we did not introduce the ICL strategy, as it would contribute additional length to the input sequence. (3) The coordinate-as-tokens approach necessitates a considerably larger-sized LLM to comprehend the numerical tokens effectively. 

\section{Results of ChartQA}\label{appendix:chartqa}

As shown in Figure~\ref{fig:failure_case_charqa}, the question-answer pairs in ChartQA~\citep{masry-etal-2022-chartqa} tend to involve the visual cues for reasoning. However, with only text and layout information as input, the proposed LayTextLLM inevitably have difficulties in reasoning visual-related information. Thus, on the ChartQA dataset, LayTextLLM can hardly achieve better performance than previous methods that include visual inputs. Although the visual information is not used in LayTextLLM, it can still exhibit better zero-shot ability than UniDoc~\citep{feng2023unidoc}. After incorporating the training set of ChartQA, the performance of LayTextLLM can be boosted to 42.2\%. Considering the importance of visual cues in ChartQA-like tasks, we will try to involve the visual information into LayTextLLM in future work. A preliminary discussion can be seen in Appendix~\ref{appendix:mllm}.

\begin{table}[h]
    \small
    \centering
    \scalebox{1.0}{
    \begin{tabular}{l|c}
    \toprule    
     ~ & ChartQA \\
    \midrule
    \textbf{OCR-free} &\\
    UniDoc~\citep{feng2023unidoc} & 10.9 \\
    DocPedia~\citep{feng2023docpedia} & 46.9$^{*}$ \\
    Monkey~\citep{li2023monkey} & 54.0$^{*}$ \\
    InternVL~\citep{chen2023internvl} & 45.6$^{*}$ \\
    InternLM-XComposer2~\citep{dong2024internlm2} & 51.6$^{*}$ \\
    TextMonkey~\citep{liu2024textmonkey} & 58.2$^{*}$\\
    TextMonkey$_{+}$~\citep{liu2024textmonkey} & \textbf{59.9}$^{*}$\\
    Qwen2-VL~\citep{wang2024qwen2} & \textbf{61.9}$^{*}$\\
    \midrule
    \textbf{Text + Coordinates} &\\
    \rowcolor[HTML]{C8E6C9}
    LayTextLLM$_{zero}$ (Ours) & 30.2\\
    \rowcolor[HTML]{C8E6C9}
    LayTextLLM$_{all}$ (Ours) & 42.6$^{*}$ \\
    \bottomrule
    \end{tabular}
    }
        \caption{Comparison with SOTA OCR-free MLLMs on ChartQA (accuracy). $^{*}$ denotes the use of the dataset's training set.}
    \label{tab:chartqa}
\end{table}

\section{Inference with SG-KIE}\label{appendix:sg_kie}

\begin{figure*}[ht]
	\centering
	\includegraphics[width=0.6\textwidth]{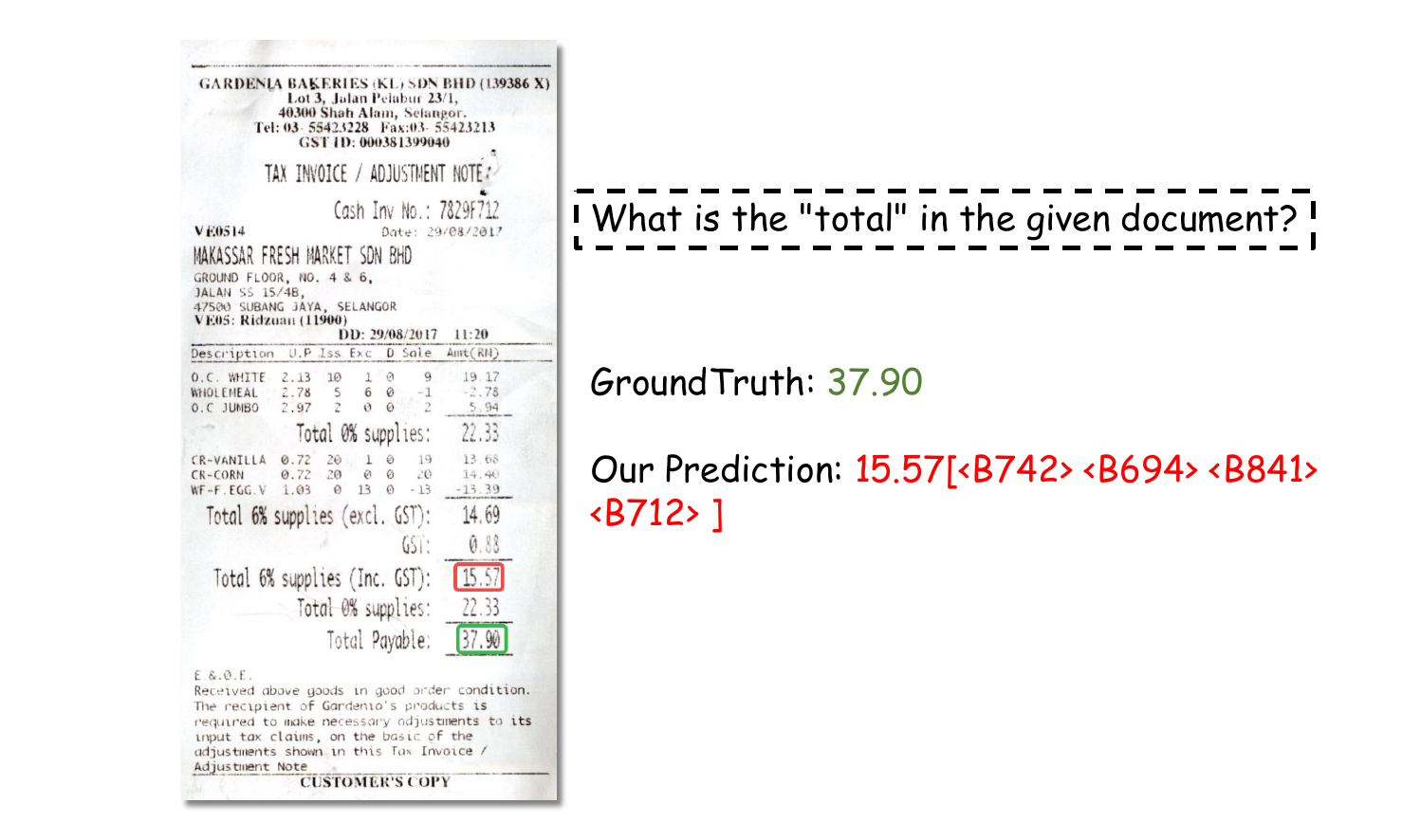}
	\caption{A failure case of SG-KIE in SROIE$^-$. The red box indicates the ground truth and the green box is the prediction.}
	\label{fig:failure_case_sroie}
\end{figure*}

As discussed in Section~\ref{sec:analysis}, incorporating SG-KIE as an auxiliary task in SFT has been shown to enhance the performance of KIE tasks. In this section, we investigate the effectiveness of using SG-KIE as a direct inference task for KIE. The results are shown in Table~\ref{tab:sg_kie_vs_normal}. We can observe that, for the FUNSD$^{-}$ and CORD$^{-}$ datasets, SG-KIE inference demonstrates improved performance. However, for the SROIE$^{-}$ dataset, there is a slight decrease in performance. We manually reviewed the problematic cases of SG-KIE and identified two main reasons for the performance drop: (1) incorrect format, which leads to parsing errors such as \textit{``432.60[ SR @ 6\%[ <B-1013><B453> <B><B478> ]''}, and (2) ambiguous key types in the SROIE$^{-}$ dataset. For instance, the key ``total'' can refer to ``grand total'' and if the model has not been trained with the dataset, SG-KIE may mistakenly localize it to the wrong value. A notable instance of this issue is shown in Figure~\ref{fig:failure_case_sroie}. These types of errors occur frequently in the dataset.

\begin{figure*}[ht]
	\centering
	\includegraphics[width=0.8\textwidth]{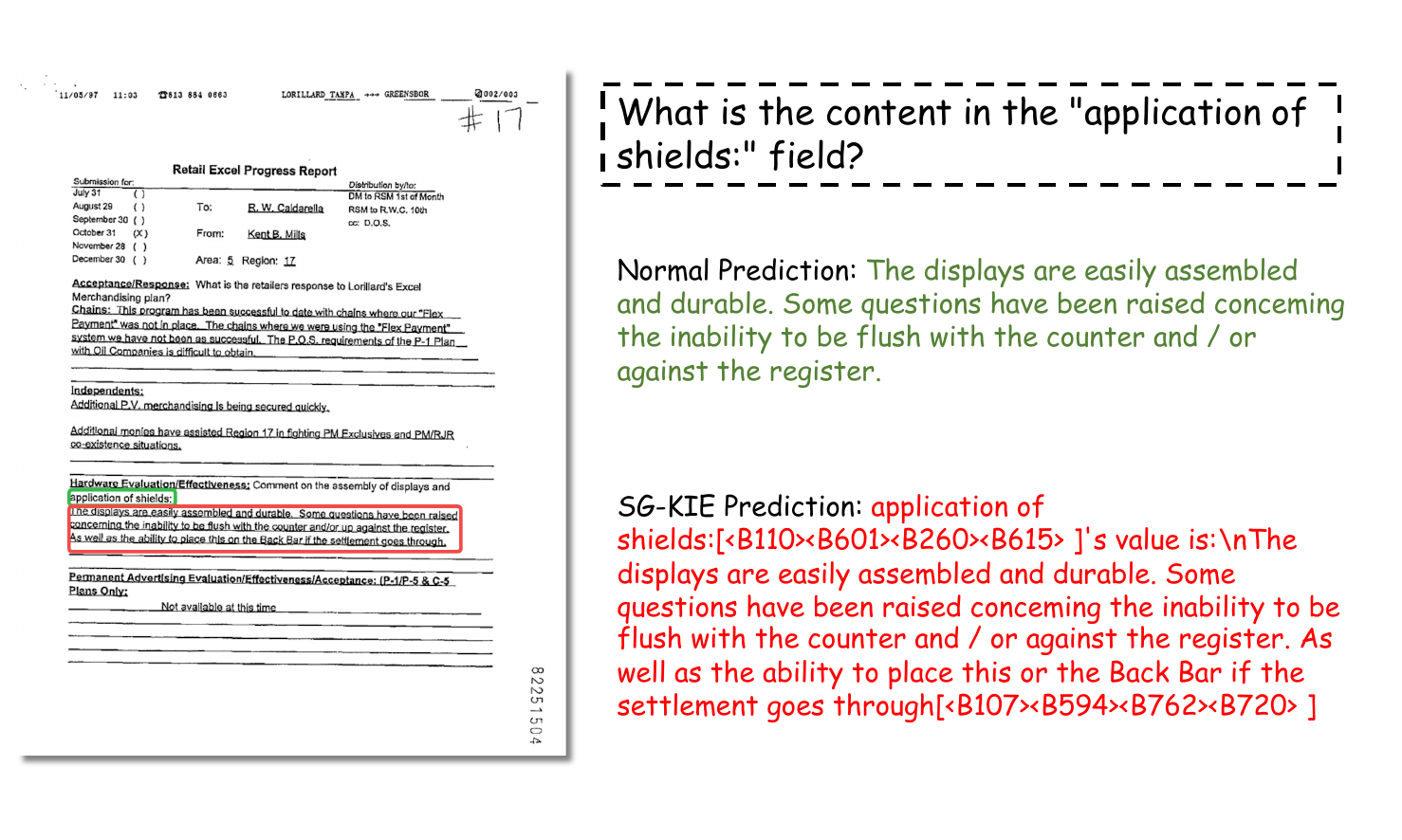}
	\caption{A good case of SG-KIE in FUNSD$^-$. The red box indicates the ground truth value and the green box is the key.}
	\label{fig:goodcase_funsd}
\end{figure*}

For improvement, we observed that SG-KIE performs better when processing complex answers that require the aggregation of multiple consecutive word-level OCR results, leading to more accurate and complete outputs, as illustrated in Figure~\ref{fig:goodcase_funsd}.

\begin{table}[h]
\centering
\scalebox{0.8}{\begin{tabular}{c|ccc}
\toprule
 Dataset & FUNSD$^{-}$  & CORD$^{-}$ & SROIE$^{-}$ \\
\midrule
 LayTextLLM$_{zero}$ &  79.6 & 81.3 & \textbf{87.0} \\
 LayTextLLM$_{zero-sg}$ & \textbf{80.0} & \textbf{81.9} & 86.0 \\
\bottomrule
\end{tabular}}
\caption{Inference with SG-KIE vs. without SG-KIE (accuracy).}\label{tab:sg_kie_vs_normal}
\end{table}

\section{Decoding Bounding Box Coordinates}\label{appendix:decode_box}

\begin{figure*}[ht]
    \centering
    \subfigure[\textit{Question}: what is the content in the "Date:" field?\newline
    \textit{Answer}: December 9, 1999]{        \vtop{\vskip0pt\hbox{\includegraphics[width=0.48\textwidth]{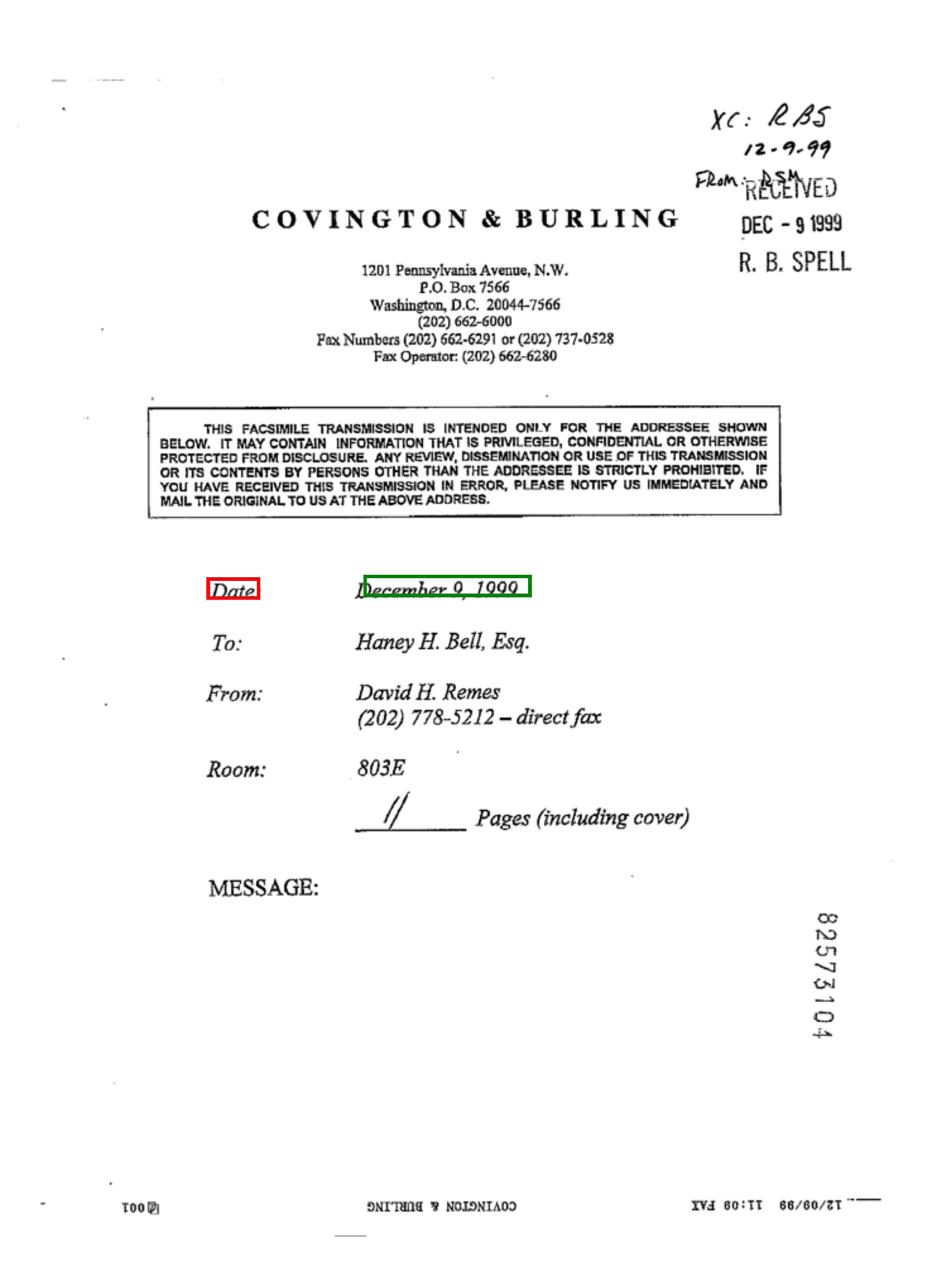}}}
        
    }
    \subfigure[\textit{Question}: what is the content in the "Pages (Including Cover)" field?\newline \textit{Answer}: 4]{        \vtop{\vskip0pt\hbox{\includegraphics[width=0.48\textwidth]{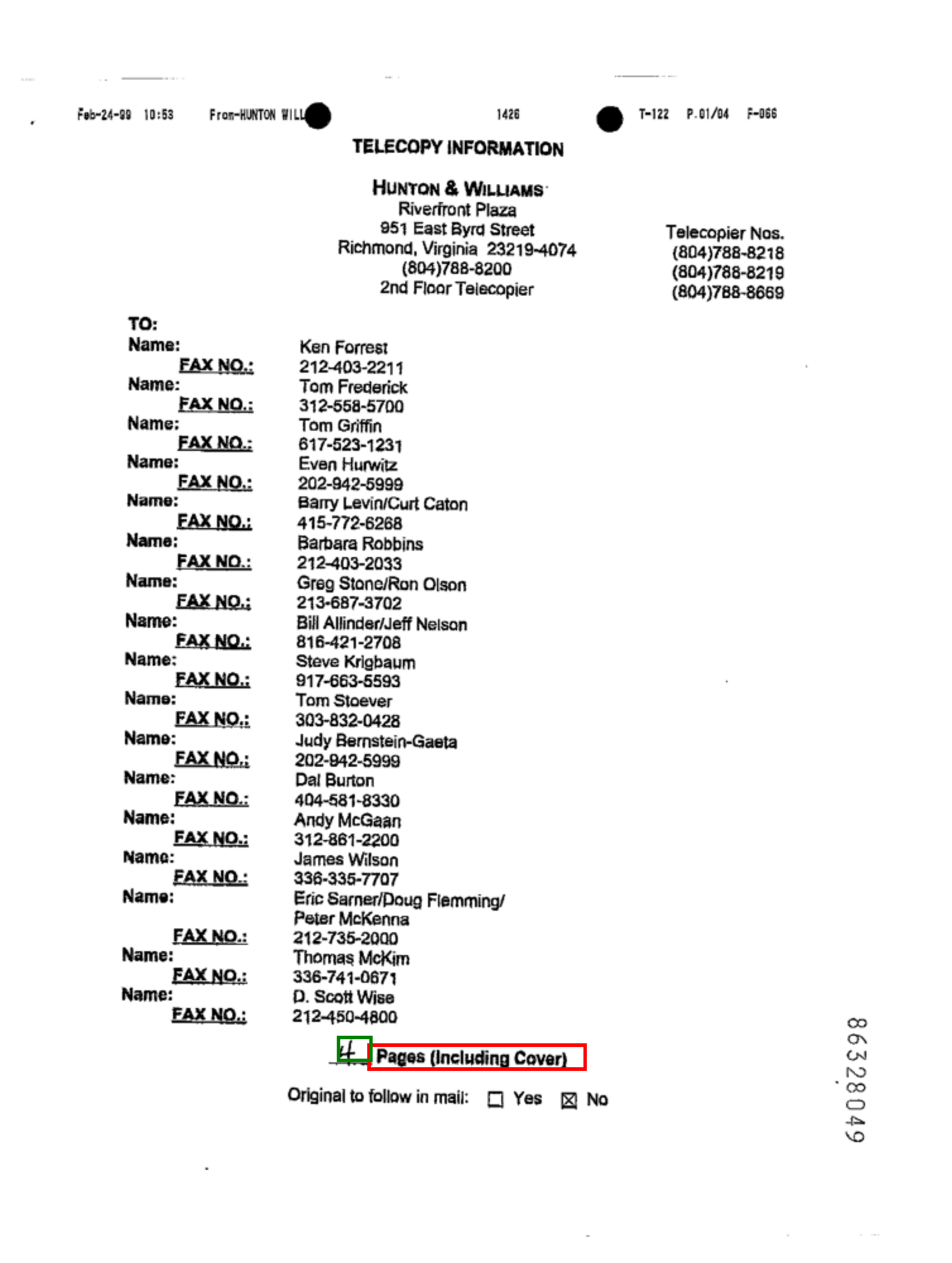}}}
       
    }
    \caption{Illustration of coordinates prediction for entity linking task. The red box indicates the key text region and the green box indicates the value text region.}
    \label{fig:box_pred_case}
\end{figure*}

\begin{figure*}[ht]
    \centering
    \subfigure[FUNSD]{        \vtop{\hbox{\includegraphics[width=0.48\textwidth]{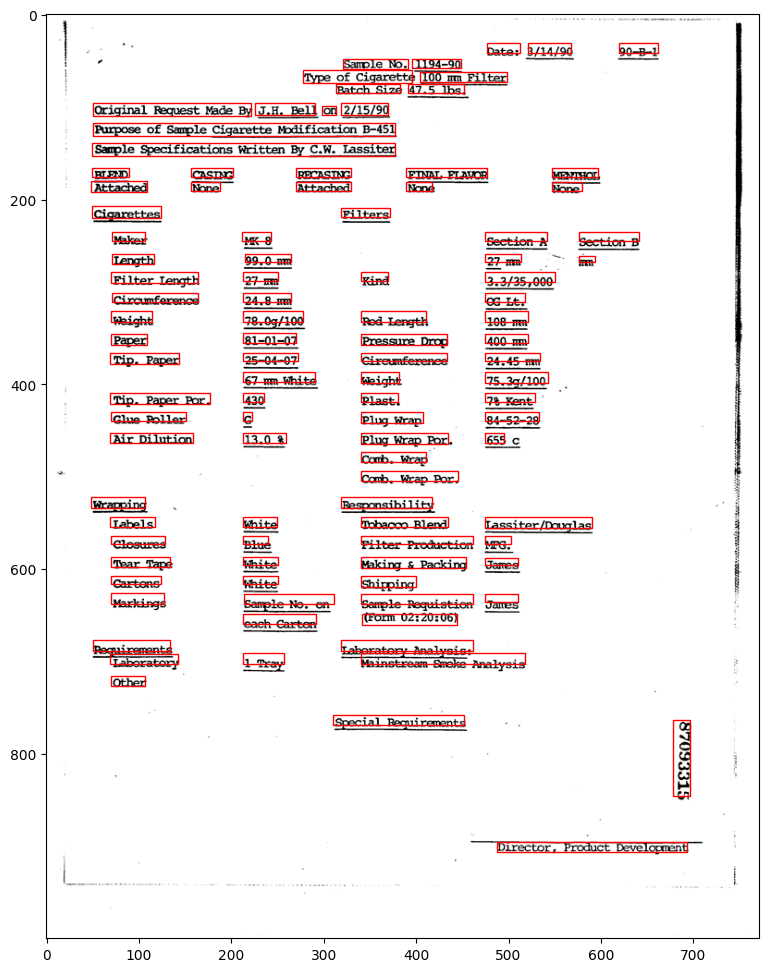}}}
        
    }
    \subfigure[FUNSD]{        \vtop{\hbox{\includegraphics[width=0.48\textwidth]{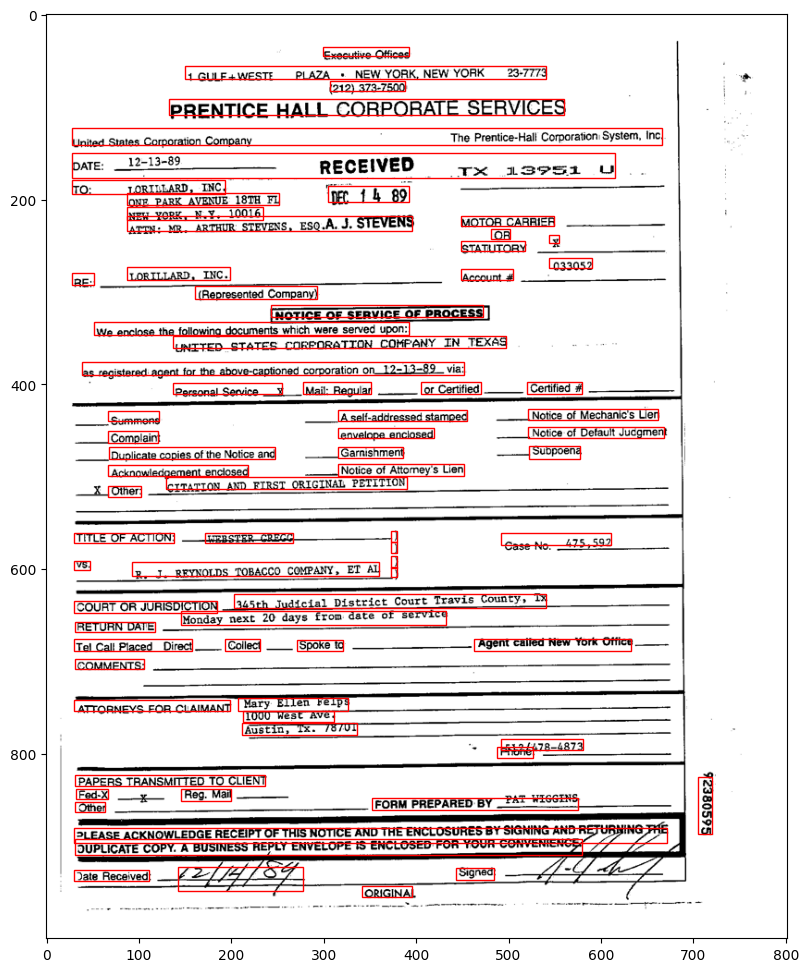}}}
       
    }
    \subfigure[POIE]{        \vtop{\hbox{\includegraphics[width=0.40\textwidth]{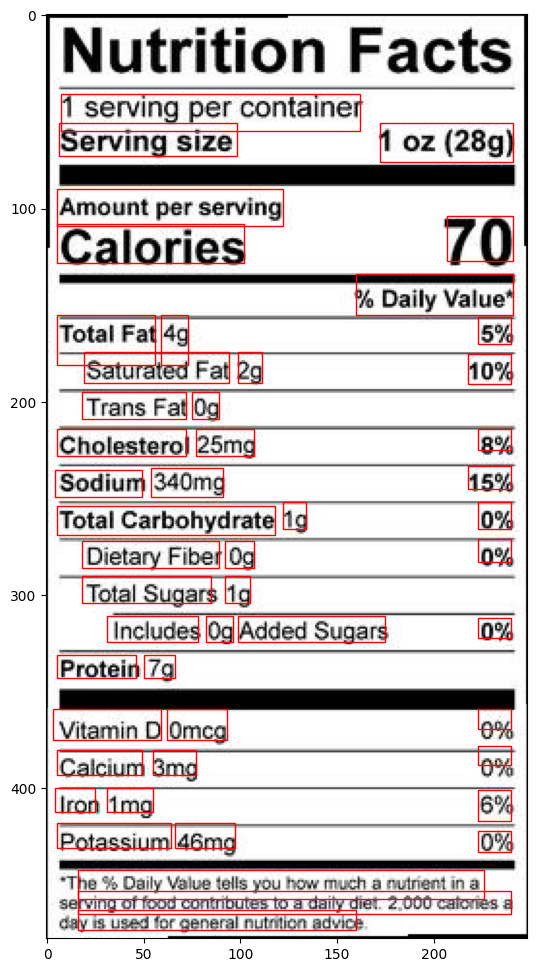}}}
       
    }
    \caption{Illustration of coordinates prediction line-level layout decoding. Documents are subsampled from OOD dataset. Red boxes are coordinates for line-level text regions.}
    
    \label{fig:line_level_case}
\end{figure*}

We also evaluate the model's ability to decode bounding box embeddings into coordinates. Since the SG-KIE task requires the model to generate precise coordinates for answers, this task can be used to assess the performance in accurately predicting bounding boxes. Specifically, we select the examples with correct predictions for textual answer and compute the Intersection over Union (IoU) score~\citep{rezatofighi2019generalized} between the predicted and ground truth coordinates. We tested the on three datasets: FUNSD, which is not used to train LayTextLLM$_{zero}$. If the IoU exceeds 0.5, we consider the bounding box prediction to be correct. Accuracy is used as the metric to evaluate this capability, we compute accuracy for the coordinates for both key and value. Results show that about 77.5\% bounding box is correctly predicted, cases are visualized in Figure~\ref{fig:box_pred_case}. Also, we visualize the coordinates prediction for the pre-training task---line-level layout decoding---in Figure~\ref{fig:line_level_case}. Moreover, SG-KIE produces coordinates, which is obviously interpretable, and providing coordinates seems to be more valuable for certain downstream tasks.

\begin{table}[h]
\centering
\scalebox{0.8}{\begin{tabular}{c|c}
\toprule
FUNSD & LayTextLLM$_{zero}$   \\
\midrule
 Accuracy &  77.5   \\
\bottomrule
\end{tabular}}
\caption{Coordinate prediction accuracy.}\label{tab:iou}
\end{table}

\section{Combination with MLLMs}\label{appendix:mllm}

As discussed in Limitation (Section~\ref{sec:limit}), LayTextLLM faces challenges with VQA tasks that require the comprehension of visual elements such as font, size, shape, objects, color, and other visual attributes. To address this limitation, we conducted a preliminary experiment combining LayTextLLM with a MLLM to explore the potential of leveraging visual information while preserving the strengths of LayTextLLM. 

Specifically, we upgrade the multimodal version of LayTextLLM by building upon Qwen2-VL and incorporating a SLP. For simplicity, neither P-LoRA nor special tokens are introduced. we layout-text alignment pre-trained and SFT the modified Qwen2-VL on the same datasets used for LayTextLLM$_{zero}$, resulting a \textit{Qwen2-VL-LayText} model. We also trained a counterpart of Qwen2-VL-LayText by incorporating only OCR text, excluding layout information. This model, which is identical in training settings to Qwen2-VL-LayText, was named \textit{Qwen2-VL-Text} and serves as a baseline. The model performance can be seen in Table~\ref{tab:qwen2_vl_laytext}. Although it shows a slight drop in performance on VQA tasks, Qwen2-VL-LayText achieves significant improvements in KIE tasks, with an overall accuracy of 76.4\% compared to 67.7\%. This further demonstrates the effectiveness of interleaving layouts and text. Interestingly, simply adding OCR text (\textit{i.e.,} Qwen2-VL-Text) also results in a notable improvement in KIE tasks when paired with Qwen2-VL. We believe this is because datasets with poor performance, such as CORD and SROIE, primarily consist of text with small or blurred fonts. In these cases, off-the-shelf OCR engines still outperform MLLMs in text recognition.

\begin{table*}[t]
    \small
    \centering
    \renewcommand\arraystretch{0.95}
    \scalebox{0.9}{
    \begin{tabular}{l|ccc|cccc}
    \toprule    
     ~ & \multicolumn{3}{c|}{\textbf{Document-Oriented VQA}} & \multicolumn{4}{c}{\textbf{KIE}} \\
     ~ & DocVQA & InfoVQA & Avg & FUNSD & CORD & SROIE & Avg \\
    \midrule
    \textbf{Metric} & \multicolumn{7}{c}{\textit{ANLS \%}} \\
    \midrule
    \textbf{Visual + Text + Coordinates} & & & & \\
    Qwen2-VL~\citep{wang2024qwen2} & \textbf{81.4} & \textbf{45.2} & \textbf{63.3} & 53.2 & 71.3 & 78.8 & 67.7 \\
    Qwen2-VL$_{text}$& 77.0 & 43.5 & 60.2 & 46.0 & 90.2 & 83.5 & 73.2 \\
    Qwen2-VL$_{LayText}$ & \textbf{81.4} & 42.7 & 62.1 & \textbf{54.2} & \textbf{91.2} & \textbf{83.7} & \textbf{76.4} \\
    \bottomrule

    \end{tabular}
    }
        \caption{Comparison with Qwen2-VL-LayText with other baselines (accuracy).}
    \label{tab:qwen2_vl_laytext}
\end{table*}

\newpage

\end{document}